\documentclass[11pt]{article}

\usepackage[final]{acl}

\usepackage{times}
\usepackage{latexsym}

\usepackage[T1]{fontenc}

\usepackage[utf8]{inputenc}

\usepackage{microtype}

\usepackage{inconsolata}

\usepackage{graphicx}
\usepackage{amsmath}
\usepackage{booktabs}
\usepackage{algorithm}
\usepackage{algorithmic}
\usepackage{float}
\usepackage{placeins}
\usepackage{booktabs}
\usepackage{multirow}
\usepackage{array}
\usepackage{makecell}
\usepackage{amsmath}
\usepackage{mathtools}
\usepackage{enumitem}
\usepackage{colortbl}
\usepackage{arydshln}
\usepackage{subcaption} 
\usepackage{graphicx} 
\usepackage{hyperref}
\usepackage{cleveref} 
\crefname{table}{Tab.}{Tabs.}
\Crefname{table}{Tab.}{Tabs.}
\crefname{section}{Sec.}{Secs.}
\Crefname{section}{Sec.}{Secs.}
\crefname{subsection}{Sec.}{Secs.}
\Crefname{subsection}{Sec.}{Secs.}
\crefname{subsubsection}{Sec.}{Secs.}
\Crefname{subsubsection}{Sec.}{Secs.}
\Crefname{figure}{Fig.}{Figs.}
\usepackage{siunitx}
\usepackage[most]{tcolorbox}
\usepackage{xcolor}
\usepackage{pifont}
\definecolor{escblue}{HTML}{2C7FB8}
\definecolor{dnorange}{HTML}{E6710E}
\definecolor{boxgray}{HTML}{F4F6F8}
\definecolor{rulegray}{HTML}{C9D2DA}

\newtcolorbox{keyfinding}[1][]{
enhanced, breakable, colback=boxgray, colframe=rulegray,
boxrule=0.4pt, left=10pt, right=8pt, top=6pt, bottom=6pt,
borderline west={2.5pt}{0pt}{escblue},
fonttitle=\bfseries, coltitle=black, title={#1}}

\newtcolorbox{playbook}[1][Practitioner playbook]{
enhanced, breakable, colback=white, colframe=escblue!85!black,
boxrule=0.8pt, arc=2pt, left=8pt, right=8pt, top=6pt, bottom=6pt,
fonttitle=\bfseries\sffamily, coltitle=white, colbacktitle=escblue!85!black,
title={#1}, attach boxed title to top left={xshift=8pt, yshift=-3pt},
boxed title style={arc=1pt, boxrule=0pt}}

\newtcolorbox{guideline}[1][]{
  enhanced,
  breakable,
  colback=green!1,
  colframe=green!5!black,
  coltitle=black,
  colbacktitle=green!10,
  fonttitle=\bfseries,
  title=#1,
  boxrule=0.5pt,
  arc=1.2pt,
  left=4pt,
  right=4pt,
  top=4pt,
  bottom=4pt,
  before skip=6pt,
  after skip=6pt
}

\usepackage{ragged2e}

\newtcolorbox{promptbox}[1]{
    enhanced,
    breakable,
    width=\linewidth,
    colback=black!2,
    colframe=black!35,
    colbacktitle=black!7,
    coltitle=black,
    boxrule=0.5pt,
    arc=1.2mm,
    left=5pt,
    right=5pt,
    top=6pt,
    bottom=6pt,
    toptitle=4pt,
    bottomtitle=4pt,
    fonttitle=\sffamily\bfseries\small,
    title={#1},
    before skip=6pt,
    after skip=8pt,
}

\usepackage{newfloat}
\usepackage{listings}
\DeclareCaptionStyle{ruled}{labelfont=normalfont,labelsep=colon,strut=off} 
\floatstyle{ruled}
\newfloat{listing}{tb}{lst}{}
\floatname{listing}{Listing}

\title{The Handoff Tax: Continuing Non-Native Trajectories in LLM Agents}

\author{
    \textbf{Roy Ganz}
    \thanks{\hspace{2pt}Equal contribution.}
    \thanks{\hspace{2pt}Correspondence: \texttt{royganz@amazon.com}.},
    \textbf{Mor Shpigel Nacson}$^{*}$,
    \textbf{Adi Kalyanpur},
    \textbf{Ron Litman}
  \\
  \\
    AWS, Agentic AI
}

\begin{document}
\maketitle

\begin{abstract}
Coding agents perform long-running tasks spanning dozens of model calls, tool uses, and code edits. As these runs unfold, users face a practical cost--quality trade-off: \emph{escalating} to a stronger model when a cheaper one struggles, or \emph{downshifting} once the hard reasoning is complete. Each switch requires the receiver to continue a \emph{non-native} trajectory produced by another model. 
We study how this handoff affects quality and cost,
and how varying the trajectory information inherited by the receiver changes the outcome.
Using pairs of low-cost, low-capability (LC) and high-cost, high-capability (HC) models from the Claude and GPT families, we vary handoff direction, timing, and interface, comparing full-trajectory transfer, compaction, and trajectory removal while preserving the repository state. Across both model families, full-trajectory escalation recovers less than half of the LC-to-HC quality gap while incurring a substantial cost premium. We term this cost--quality penalty the \emph{handoff tax}. By contrast, downshift offers a favorable cost--quality point. Interestingly, the preferred interface also reverses with direction: reducing LC-model trajectory information improves escalation quality, whereas removing the HC-model trajectory reduces downshift quality.
\end{abstract}


\section{Introduction}
\label{sec:intro}

\begin{figure}[t]
\centering
\includegraphics[width=\linewidth]{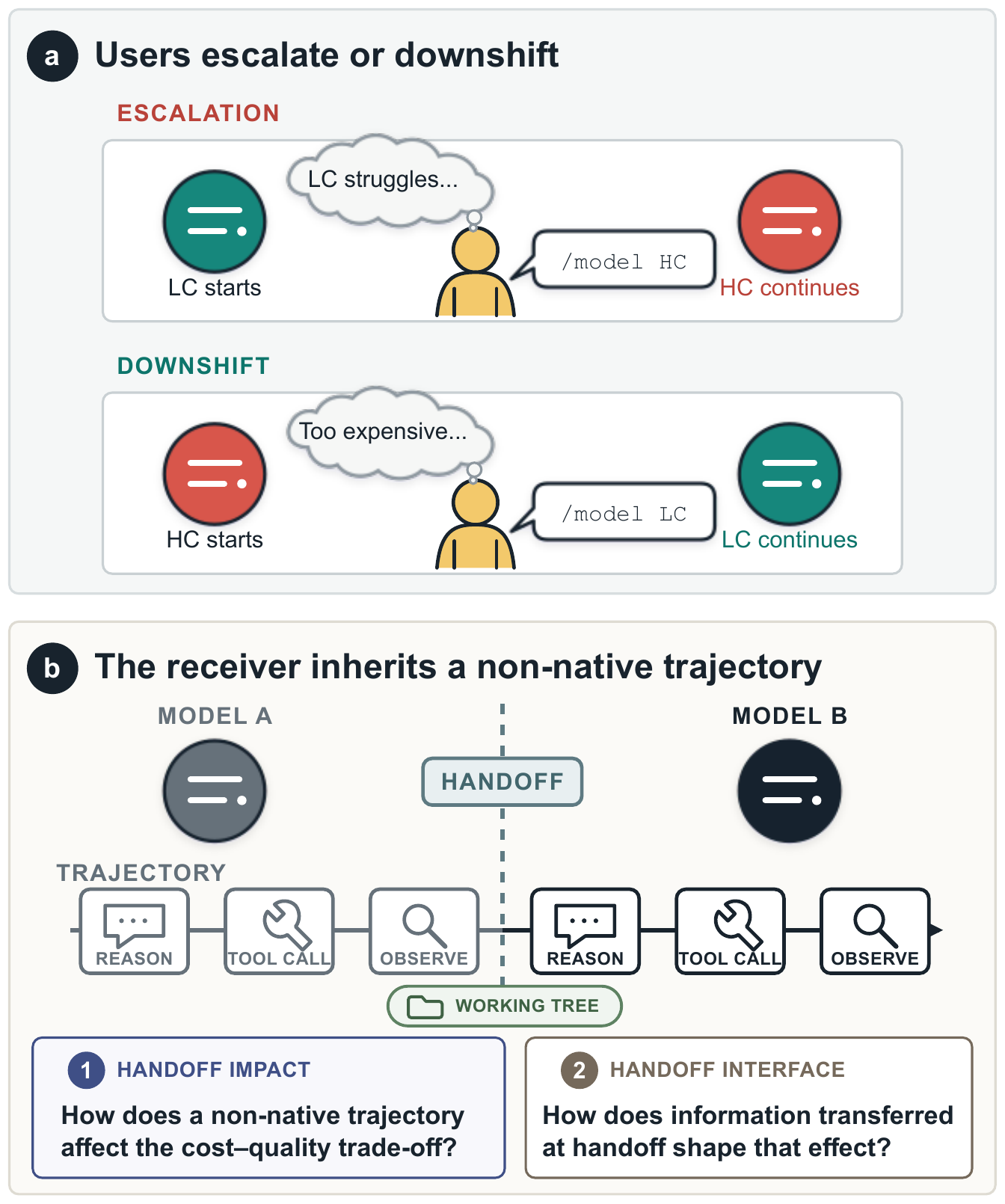}
\caption{\textbf{The Handoff Tax.}
\textbf{(a)}~Users escalate to a high cost model (HC) when the cheaper model (LC)
struggles, or downshift to a cheaper one once hard reasoning appears complete.
\textbf{(b)}~At handoff, the receiver inherits another model's trajectory.
We ask how this affects cost and quality, and how varying the transferred
trajectory information shapes the effect.
}
\label{fig:model-handoff}
\end{figure}

Coding agents now routinely chain dozens to hundreds of model calls on a single task, making model choice an economic decision as much as a technical one~\cite{yao2023reactsynergizingreasoningacting,schick2023toolformerlanguagemodelsteach,qin2023toolllmfacilitatinglargelanguage,patil2023gorillalargelanguagemodel,pmlr-v235-wang24h,wang2025openhandsopenplatformai,xu2025theagentcompanybenchmarkingllmagents,hitzig2026agentic}. 
Available models span distinct cost--quality profiles: higher-capability models resolve more issues but are substantially more expensive, whereas cheaper models offer lower cost at reduced capability. Users must often choose among these models before knowing what level of capability the task actually requires.
Model switching offers a natural response and is already supported directly by coding-agent products. Users may start with a low-capability model and \emph{escalate} to a stronger one when the agent struggles; conversely, once the hard reasoning appears complete, they may \emph{downshift} to a cheaper model to finish the task at lower cost. Throughout, \textbf{LC} and \textbf{HC} denote the lower-cost/lower-capability and higher-cost/higher-capability models within a family; in our experiments, these correspond to Claude and GPT model pairs. Despite being exposed in several coding-agent interfaces (e.g., \texttt{/model} commands in Kiro, Codex, and Claude Code), the cost--quality consequences of capability handoffs in long-horizon coding agents remain poorly understood.

A mid-task handoff places a model in an unusual position: it must continue a long trajectory that \emph{another} model produced. Unlike ordinary generation, where models extend their own trajectory--their own phrasing, hypotheses, tool-use idioms, and dead ends--a handoff requires them to inherit a trajectory they did not create. This inherited context may contain reasoning the receiver would not have generated and mistakes it would not have made. Whether it helps or hurts may depend on direction:
an HC model might be \emph{anchored} by LC's wrong turns, while
an LC model might coast on HC's groundwork--or flounder when extending
reasoning beyond its own ability.

\begin{figure*}[t]
\centering
\includegraphics[width=\textwidth]{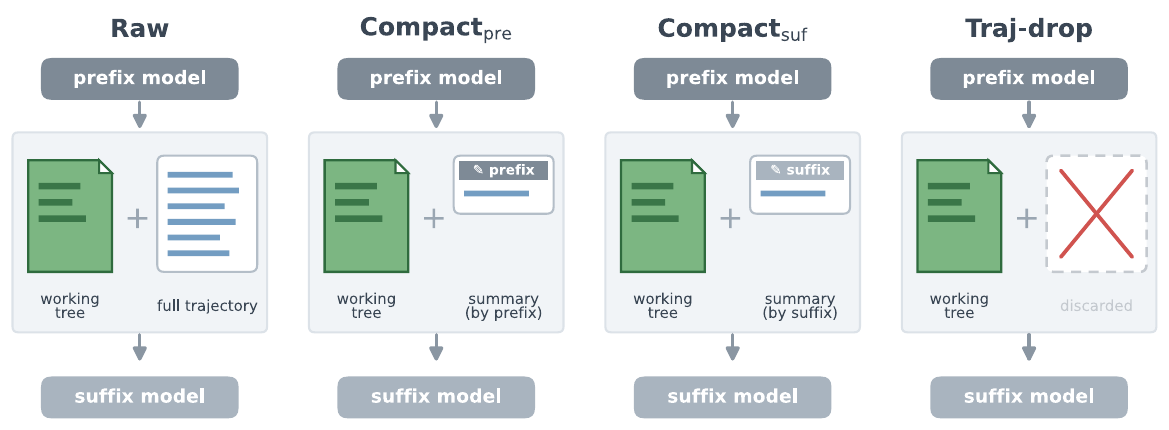}
\caption{
\textbf{Handoff interfaces.} All strategies preserve the prefix model's edited working tree at the switch but vary the trajectory information transferred to
the suffix: the full trajectory (\textbf{Raw}), a summary written by the prefix or suffix model (\textbf{Compact$_{\mathrm{pre}}$}/\textbf{Compact$_{\mathrm{suf}}$}), or none (\textbf{Traj-drop}).}
\label{fig:methods}
\end{figure*}

We therefore ask two questions: whether and how inheriting a non-native
trajectory changes the cost--quality trade-off, and how the information
transferred at the handoff shapes that effect. We study these questions
systematically on SWE-bench Verified~\cite{jimenez2024swebench} along three
axes: the \emph{direction} of the switch (LC$\to$HC and HC$\to$LC), its \emph{timing} (swept across difficulty-calibrated percentiles),
and the \emph{interface}, which determines what trajectory information the
receiver inherits. The standard \emph{raw handoff} passes the full trajectory
verbatim. Alternatively, the trajectory can be compacted into a summary written
by either the outgoing or incoming model, or dropped while preserving the code
edits already written to disk (\emph{Traj-drop}). \Cref{fig:methods}
illustrates these interfaces.

\paragraph{Findings and contributions.}
We present, to our knowledge, the first systematic study of mid-trajectory
model capability handoffs in long-horizon coding agents. Across two model
families, both handoff directions, seven switch points, and four interfaces,
our primary study comprises \textbf{58 configurations per family,
58{,}000 runs, and 36 billion tokens processed} overall. Three main findings emerge
from the coding-agent study:




\paragraph{1. Raw escalation recovers limited quality and can be dominated by
restarting with HC.}
Raw continuation, the default escalation interface, recovers less than half of
HC's quality advantage in both model families. Most strikingly, for Claude,
even after paying for the LC prefix, abandoning the attempt and restarting
with HC is cheaper and more accurate than Raw continuation.

\paragraph{2. Continuing another model's trajectory has direction-dependent
value.}
Unlike escalation, Raw downshift provides a favorable cost--quality trade-off:
LC receivers retain substantial quality gains
over LC while remaining cheaper than HC. The balance differs across
families: Claude retains most of LC's cost advantage, whereas GPT retains most
of HC's quality advantage.

\paragraph{3. Handoff interfaces reveal what trajectory information helps the
receiver.}
In escalation, reducing LC trajectory information improves the
trade-off: compaction favors savings, whereas trajectory removal favors quality
recovery. In downshift, removing the HC trajectory while preserving
its working-tree edits reduces quality across both model families. This reveals
a directional duality: HC trajectories guide LC receivers, whereas
LC trajectories burden HC receivers.

\section{Related Work}
\label{sec:related}

Prior work improves the cost--quality trade-off of LLM inference through model
selection. Cascades invoke higher-capability models when lower-cost ones are insufficient,
while routers dispatch requests over a model pool
\citep{chen2023frugalgptuselargelanguage,yue2024largelanguagemodelcascades,hu2024routerbenchbenchmarkmultillmrouting,ding2024hybridllmcostefficientqualityaware,aggarwal2025automixautomaticallymixinglanguage,
valkanas2025c3pooptimizedlargelanguage,ong2025routellmlearningroutellms,kotte2026uccicalibrateduncertaintycostoptimal}. Recent work extends routing to multi-turn
interactions, deciding which model acts and when
\citep{zhang2026myopicselectionlonghorizonawareness,zhang2026mtroutercostawaremultiturnllm,
hemadri2026r2vagent}. SWE-Router~\citep{son2026swerouter} uses an LC model's
partial coding trajectory to decide whether it should continue or an HC
model should restart, but does not transfer the trajectory.
\citet{khraishi2026performancedrift} study directional drift when one model
continues another's dialogue prefix, but only at the final turn and without
alternative handoff interfaces.
\citet{kc2026handoffdebt} study agents taking over interrupted coding tasks, examining how the information exposed from the prior run affects the effort needed to rediscover the predecessor's work.
Their models are evaluated as successors
rather than as LC--HC alternatives with distinct cost profiles.

Our work instead studies a practical user pattern: switching during an active
run between commercial models with different capability and cost profiles. We
vary direction, timing, and interface to measure how continuing a
\emph{non-native} trajectory affects end-to-end quality and monetary cost.

\section{Experimental Framework}
\label{sec:framework}



We study \emph{mid-trajectory model handoffs} in long-horizon coding tasks,
using SWE-bench Verified~\cite{jimenez2024swebench}, a real GitHub issue-resolution benchmark with executable
pass/fail evaluation and difficulty annotations. All conditions use the same
\texttt{mini-swe-agent}\footnote{https://github.com/swe-agent/mini-swe-agent} scaffold, tools, and prompts~\cite{yang2024sweagentagentcomputerinterfacesenable}. Within each family,
\textbf{LC} denotes the lower-cost, lower-capability model and \textbf{HC} the
higher-cost, higher-capability model: Haiku~4.5~\cite{anthropic2025claudehaiku45} and Opus~4.7~\cite{anthropic2026claudeopus47} for Claude, and
GPT-5.6 Luna and Sol for GPT~\cite{openai2026gpt56}. 

\subsection{Handoff Conditions}
\paragraph{Directions.}
We evaluate the resulting cost--quality trade-off in both model switch directions.
\emph{Escalation} (LC$\rightarrow$HC) starts with LC and switches to HC. It
tests whether HC can rescue a trajectory begun by LC and in what cost. \emph{Downshift}
(HC$\rightarrow$LC) starts with HC and transfers the remaining work to LC. It
tests how much of HC's quality advantage remains and how it affects the cost.

\paragraph{Handoff Strategies.}
The central design variable is \emph{what crosses the model boundary}. Let
$\mathcal{T}_{1:K}$ denote the trajectory produced by the prefix model up to
step $K$, and $\mathcal{W}_K$ the working tree (edited files on disk) at that
point. All strategies preserve $\mathcal{W}_K$; they differ only in the
trajectory information passed to the suffix model
(Figure~\ref{fig:methods}):

\begin{itemize}[leftmargin=*,itemsep=0pt]
\item \emph{Raw}: the full trajectory $\mathcal{T}_{1:K}$ is transferred.
\item \emph{Compact$_{\mathrm{pre}}$}: the prefix model compacts
      $\mathcal{T}_{1:K}$; only the summary is passed.
\item \emph{Compact$_{\mathrm{suf}}$}: the suffix model compacts
      $\mathcal{T}_{1:K}$ and continues from
      that summary alone.
\item \emph{Traj-drop}: no trajectory information is transferred; the suffix
      model begins with only $\mathcal{W}_K$.
\end{itemize}

\noindent For escalation, we construct two restart controls that charge for LC's
work but discard both its trajectory and working-tree edits before restarting
HC from the original task state.
\emph{Abort + HC fresh} charges for LC through
step $K$, whereas \emph{LC-full + HC-full} charges for a complete LC run. These
test whether \emph{any} continuation of the LC trajectory is preferable to
abandoning it outright.

\subsection{Switch Points}
\label{sec:switchpoints}

A handoff's effect depends on \emph{when} it occurs. Fixed step counts are not
comparable across tasks or models because trajectory lengths vary. We therefore
define switch points as percentiles of the starting model's step-count
distribution. Computing these percentiles over the full dataset would skew the
difficulty distribution at each switch point relative to the full dataset,
since easier tasks tend to finish earlier than harder ones. To avoid this bias,
we estimate each percentile separately within each task-difficulty bucket.
For example, $\mathrm{p}25$ is the 25th-percentile step count for the
relevant starting model and difficulty bucket.
We sweep $K$ over percentiles $\{5,10,15,25,35,45,50\}$.
The main results average over all
seven switch points. 
We omit later
percentiles because too few instances reach the switch for reliable
comparison. Appendix~\Cref{tab:step-distribution}
report the exact bucket-specific steps.

\subsection{Evaluation on the Switched Subset}
\label{sec:switched-subset}


Comparing handoffs fairly requires accounting for which runs actually underwent a handoff. If the starting model finishes before step $K$, no switch occurs; the run is a single-model trajectory and provides no evidence about the handoff effect. We therefore evaluate the \textbf{switched subset}: for each direction and $K$, we restrict comparisons to the intersection of instances that switched under all four strategies. This ensures that every strategy is evaluated on the same tasks that underwent a handoff. We evaluate the baselines on this same subset to provide matched quality and cost references.

\subsection{Metrics}
\label{sec:metrics}

We report raw pass rate, mean cost (USD), and mean step count. Costs follow
provider pricing, including cache reads and writes. We compute these metrics
separately at each switch point and average them within each reported window.

\paragraph{Normalized metrics.}
%
Following \Cref{sec:switched-subset}, we evaluate each strategy at each switch point relative to the LC-only and HC-only baselines on its matched switched subset, placing all switch points on the same LC-to-HC reference scale.
Let
$R_m(K)$ and $C_m(K)$ denote pass rate and mean cost for strategy $m$; LC/HC
subscripts denote the matched single-model baselines. \textbf{Quality Recovery
(QRec)} is the fraction of HC's quality advantage over LC that the strategy
recovers, while \textbf{Cost-Savings Retention (CSRet)} is the fraction of
LC's cost advantage over HC that it retains:
\begin{align}
\mathrm{QRec}(m,K)
&=100\frac{R_m(K)-R_{\mathrm{LC}}(K)}
             {R_{\mathrm{HC}}(K)-R_{\mathrm{LC}}(K)},
\label{eq:qrec}
\\
\mathrm{CSRet}(m,K)
&=100\frac{C_{\mathrm{HC}}(K)-C_m(K)}
             {C_{\mathrm{HC}}(K)-C_{\mathrm{LC}}(K)}.
\label{eq:csret}
\end{align}
QRec is $0$ at LC quality and $100$ at HC quality; CSRet is $100$ at LC cost
and $0$ at HC cost. Values outside $[0,100]$ fall beyond these
anchors (\textit{e.g.}, negative CSRet is costlier than HC-only).

\paragraph{Experimental scale.}
For each model family, we evaluate 58 configurations: two single-model
baselines and 56 handoff configurations spanning seven switch points, four
handoff strategies, and two directions. Running all configurations on the 500
SWE-bench Verified instances across both model families yields 58{,}000 agent
runs, 2 million LLM API calls, and 36 billion processed tokens.


\section{The Handoff Tax}
\label{sec:handoff}

\begin{table*}[t]
\centering
\small
\setlength{\tabcolsep}{2.6pt}
\renewcommand{\arraystretch}{1.04}
\resizebox{\textwidth}{!}{%
  \begin{tabular}{@{}lrrrrrlrrrrr@{}}
    \toprule
    \multicolumn{6}{c}{\textbf{Claude} (Haiku~4.5 / Opus~4.7)} &
    \multicolumn{6}{c}{\textbf{GPT-5.6} (Luna / Sol)} \\
    \cmidrule(lr){1-6}\cmidrule(lr){7-12}
    Strategy & Pass & Cost & Steps & QRec & CSRet &
    \hspace{9pt}Strategy & Pass & Cost & Steps & QRec & CSRet \\
    \midrule
    \multicolumn{12}{l}{\textbf{(a) Escalation (LC~$\rightarrow$~HC)}} \\
    \rowcolor{gray!12} LC-only & 60.7 & 0.40 & 80 & $0$ & $100$ & \hspace{9pt}LC-only & 58.7 & 0.06 & 14 & $0$ & $100$ \\
    \rowcolor{gray!12} HC-only & 79.2 & 0.72 & 31 & $100$ & $0$ & \hspace{9pt}HC-only & 83.7 & 0.47 & 17 & $100$ & $0$ \\
    \noalign{\vskip 2pt}
    \cdashline{1-12}
    \noalign{\vskip 2pt}
    \rowcolor{gray!12} Abort + HC fresh & 79.2$^\dagger$ & 0.90 & 78 & $100^\dagger$ & $-58$ & \hspace{9pt}Abort + HC fresh & 83.7$^\dagger$ & 0.51 & 27 & $100^\dagger$ & $-11$ \\
    \rowcolor{gray!12} LC-full + HC-full & 79.2$^\dagger$ & 1.12 & 111 & $100^\dagger$ & $-130$ & \hspace{9pt}LC-full + HC-full & 83.7$^\dagger$ & 0.53 & 31 & $100^\dagger$ & $-15$ \\
    \midrule
    Raw handoff & 69.2 & 1.61 & 74 & $47$ & $-285$ & \hspace{9pt}Raw handoff & 67.5 & 0.36 & \textbf{17} & $36$ & $26$ \\
    Compact$_{\text{pre}}$ & 71.8 & \textbf{0.75} & 71 & $60$ & $\mathbf{-11}$ & \hspace{9pt}Compact$_{\text{pre}}$ & 68.8 & \textbf{0.27} & 18 & $40$ & $\mathbf{49}$ \\
    Compact$_{\text{suf}}$ & 69.6 & 0.98 & \textbf{67} & $49$ & $-82$ & \hspace{9pt}Compact$_{\text{suf}}$ & 68.8 & 0.43 & 19 & $40$ & $10$ \\
    Traj-drop & \textbf{72.4} & 0.81 & 74 & $\mathbf{64}$ & $-30$ & \hspace{9pt}Traj-drop & \textbf{79.7} & 0.50 & 25 & $\mathbf{84}$ & $-8$ \\
    \specialrule{\lightrulewidth}{3pt}{1pt}
    \specialrule{\lightrulewidth}{0pt}{3pt}
    \multicolumn{12}{l}{\textbf{(b) Downshift (HC~$\rightarrow$~LC)}} \\
    \rowcolor{gray!12} LC-only & 54.6 & 0.41 & 79 & $0$ & $100$ & \hspace{9pt}LC-only & 63.6 & 0.05 & 13 & $0$ & $100$ \\
    \rowcolor{gray!12} HC-only & 75.8 & 0.85 & 35 & $100$ & $0$ & \hspace{9pt}HC-only & 85.8 & 0.47 & 18 & $100$ & $0$ \\
    \midrule
    Raw handoff & 65.6 & \textbf{0.51} & 58 & $50$ & $\mathbf{80}$ & \hspace{9pt}Raw handoff & \textbf{81.0} & \textbf{0.41} & \textbf{16} & $\mathbf{79}$ & $\mathbf{14}$ \\
    Compact$_{\text{pre}}$ & \textbf{66.8} & 0.52 & \textbf{54} & $\mathbf{56}$ & $78$ & \hspace{9pt}Compact$_{\text{pre}}$ & 79.8 & 0.43 & 18 & $72$ & $10$ \\
    Compact$_{\text{suf}}$ & 63.7 & 0.53 & 65 & $42$ & $73$ & \hspace{9pt}Compact$_{\text{suf}}$ & 80.4 & 0.42 & 17 & $75$ & $13$ \\
    Traj-drop & 60.9 & 0.59 & 79 & $28$ & $59$ & \hspace{9pt}Traj-drop & 75.4 & 0.43 & 24 & $53$ & $10$ \\
    \bottomrule
  \end{tabular}
}
\caption{\textbf{Aggregate coding-agent handoffs across model pairs.}
Raw escalation recovers less than half of HC's quality advantage, whereas
downshift offers a favorable intermediate cost--quality point.
Reducing inherited LC context improves escalation, while preserving
HC context improves downshift.
Values are unweighted means over seven switch points on matched switched
subsets; QRec and CSRet follow~\Cref{sec:metrics}.
}
\label{tab:handoff_duality}
\end{table*}

\begin{figure*}[t]
\centering
\includegraphics[width=\textwidth]{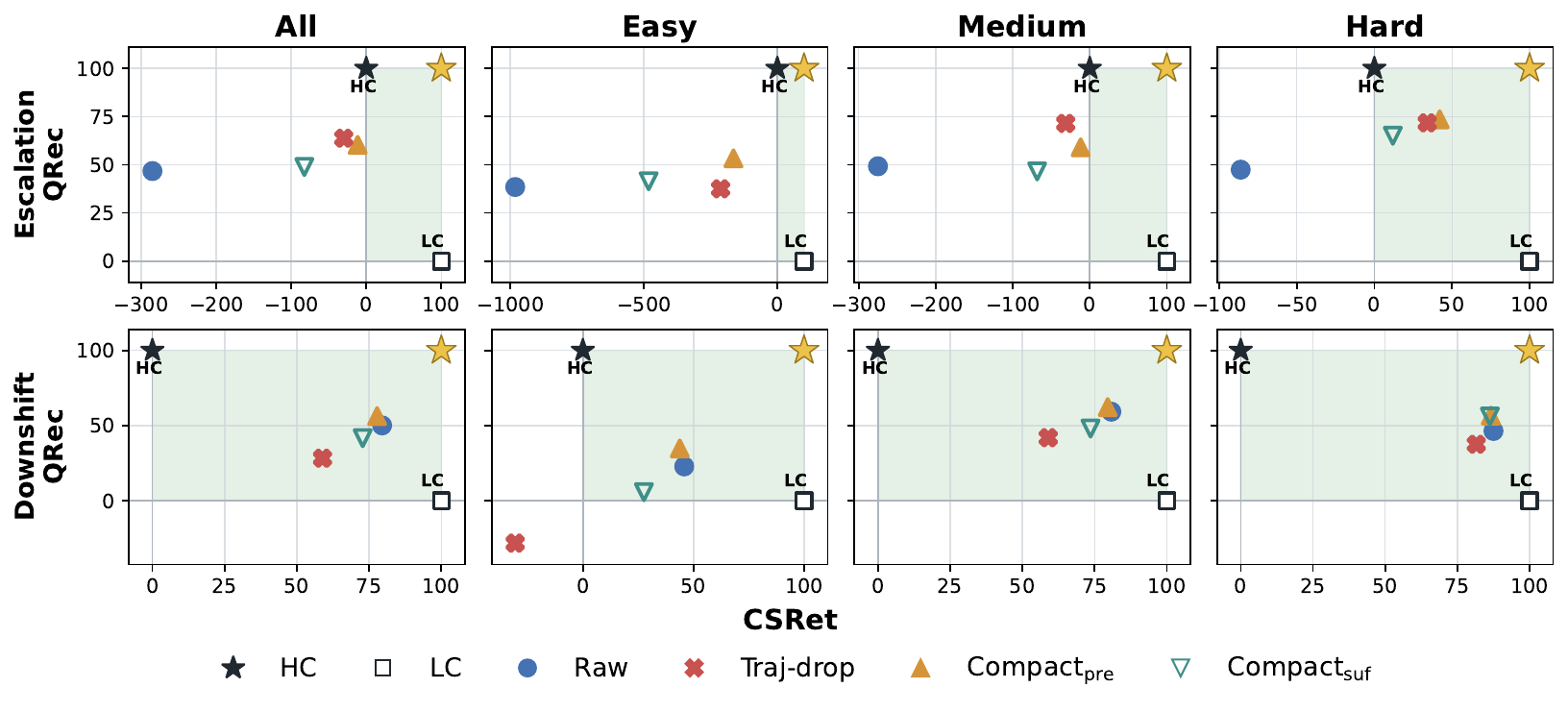}
\caption{\textbf{Handoff effects vary with task difficulty.}
For Claude, all escalation interfaces are unfavorable on easy and medium tasks, but on hard tasks reduced-context interfaces become cheaper than HC-only while recovering 65--74\% of HC's quality advantage; Raw does not make this transition.
In downshift, dropping the HC trajectory yields the worst cost--quality trade-off among tested interfaces in every difficulty
bucket.}
\label{fig:practitioner-difficulty}
\end{figure*}

We evaluate both switching directions under the protocol of \Cref{sec:framework} and report matched switched-subset results in \Cref{tab:handoff_duality}.
Across both model families, two qualitative patterns emerge.
First, Raw handoff recovers less than half of HC's quality advantage in escalation, but offers a favorable cost--quality trade-off in downshift at lower cost than HC-only.
Second, inherited trajectory has direction-dependent value: dropping LC trajectory improves escalation, whereas preserving HC trajectory improves downshift.
These quality patterns recur across families, while their cost implications differ because the model pairs have different baseline trajectory lengths and HC-to-LC per-task cost ratios.
We examine escalation (\Cref{sec:escalation}), downshift
(\Cref{sec:downshift}), and the receiver computations underlying these
effects (\Cref{sec:mechanics}).
\Cref{app:breakdowns_swe} reports uncertainty estimates for the interface
comparisons.

\subsection{Escalation: Limited Rescue at High Cost}
\label{sec:escalation}

\paragraph{Raw escalation fails to recover most of HC's quality advantage.}
Raw is the default handoff interface, passing the full LC trajectory directly to HC.
Across both model pairs in Panel~(a) of \Cref{tab:handoff_duality}, it recovers
less than half of the LC--HC quality gap (QRec$=47\%$ for Claude and 36\% for
GPT). This limited quality recovery is common to both families; whether it also
creates a severe cost premium depends on the model pair.

\paragraph{Raw's limited rescue comes at substantial cost.}
Raw escalation substantially increases cost over LC-only in both families, by
approximately $4.0\times$ for Claude and $6.1\times$ for GPT. For Claude, this
increase is large enough that Raw costs more than twice as much as starting
with HC (\$1.61 vs.\ \$0.72). By contrast, Raw remains cheaper than HC-only
for GPT (\$0.36 vs.\ \$0.47). This difference reflects the model pairs'
baseline per-task economics: HC-only costs roughly $8\times$ as much as LC-only for
GPT, compared with roughly $2\times$ for Claude.

For Claude, Raw's limited rescue and severe cost premium motivate a stricter
question: once LC work has been incurred, is it better to continue Raw or
abandon the run and restart HC? We compare Raw with two controls that charge
for LC's work but discard both its conversation and working-tree edits before
restarting HC from the original task state. \emph{Abort + HC fresh} charges for
LC through the switch point before running HC from scratch;
\emph{LC-full + HC-full} charges for a complete LC run before restarting HC.
Both recover HC-only quality by construction. Even
after charging for the discarded LC work, both controls remain cheaper than
Raw (\$0.90 and \$1.12, respectively, versus \$1.61). Thus, for the Claude
pair, Raw escalation is strictly dominated by restarting HC from scratch:
restarting costs less and solves more tasks.

We next test whether reducing the inherited trajectory can improve Raw's
limited rescue without discarding LC's working-tree edits.

\paragraph{Different trajectory reductions favor different cost--quality objectives.}
Both compaction interfaces yield slightly higher observed pass rates than Raw,
with differences of 0.4--2.6 points (\Cref{tab:handoff_duality}).
Compact$_{\mathrm{pre}}$ provides the highest cost retention among the tested
escalation interfaces in both families. For Claude, it raises QRec from 47\%
to 60\% and improves CSRet from $-285\%$ to $-11\%$, approaching HC-only cost
parity. For GPT, QRec changes only slightly, from 36\% to 40\%, while CSRet
improves from 26\% to 49\%.

Taking context reduction further produces a larger quality gain. Traj-drop
removes the trajectory while preserving LC's working-tree edits. With
Traj-drop, QRec rises from 47\% to 64\% for Claude and from 36\% to 84\% for
GPT. This quality recovery comes with negative CSRet in both families
($-30\%$ and $-8\%$), meaning that Traj-drop remains costlier than HC-only.

\paragraph{Difficulty changes when Claude escalation is worthwhile.}
In aggregate, every Claude escalation interface costs more than HC-only,
suggesting that escalation may never be worthwhile for this pair. We test
whether this conclusion holds across task difficulty using the SWE-bench
Verified labels. \Cref{fig:practitioner-difficulty} reveals an exception on
hard tasks.

On easy tasks, all interfaces offer a poor cost--quality
trade-off, recovering only around half of the LC--HC quality gap while costing
far more than HC-only (CSRet $-164\%$ to $-982\%$; \Cref{tab:difficulty_handoff}).
In the hard-task bucket, by contrast, all three reduced-context
interfaces cost less than HC-only, retaining 12--42\% of LC's cost
advantage while recovering 65--74\% of the LC--HC quality gap. Raw does not
make the same transition: it remains costlier than HC-only and recovers only
47\% of the quality gap. Thus, difficulty alone does not make Raw escalation cost-effective,
but it can make reduced-context escalation attractive. Because the hard
subset is small ($\bar{N}\!\approx\!24$ per cell), we treat this pattern as
exploratory rather than as a cross-family finding.

\paragraph{Takeaway.}
Raw escalation is a poor bargain: it recovers less than half of HC's quality
advantage at several times LC’s cost. For Claude, it even
costs more than completing LC and then running HC from scratch. What HC
inherits changes the trade-off: Compact$_{\mathrm{pre}}$ minimizes handoff
cost, while Traj-drop maximizes quality recovery.

\subsection{Downshift: Context Preserves Quality}
\label{sec:downshift}

\paragraph{Raw downshift lets LC build on HC's work.}
In contrast to the limited rescue produced by Raw escalation, the same
interface transfers useful HC progress when the direction reverses. Downshift
tests whether LC can build on HC's partial work.
For Claude, Raw HC$\rightarrow$LC raises pass rate from 54.6\% to 65.6\%,
while cost rises only from \$0.41 to \$0.51, retaining 80\% of LC's cost
advantage. GPT shows a different balance: Raw retains 79\% of HC's quality
advantage but only 14\% of LC's cost advantage. Thus, both LC receivers
convert HC's partial work into a favorable intermediate point below HC-only
cost, with different balances: Claude retains most of LC's cost advantage,
whereas GPT retains most of HC's quality advantage.

\paragraph{Removing the HC trajectory harms downshift quality.}
Raw downshift transfers both HC's working-tree edits and its trajectory, so
its quality gain does not reveal which form of state helps LC. 
Traj-drop isolates the role of trajectory guidance by preserving the edits while removing the trajectory.
For Claude, Traj-drop---the highest-quality escalation interface---becomes the
lowest-quality and most expensive tested downshift strategy, recovering only 28\% of
the LC--HC quality gap, compared with 50\% for Raw and 56\% for
Compact$_{\mathrm{pre}}$, while retaining just 59\% of LC's cost advantage.
The difficulty breakdown in \Cref{fig:practitioner-difficulty} shows that Traj-drop has the same quality-and-savings disadvantage across all
difficulty levels.

The same quality reversal appears for GPT. Traj-drop recovers only 53\% of the
LC--HC quality gap, compared with 79\% for Raw and 72--75\% for the two
compaction interfaces. Cost retention spans only 10--14\% across the GPT
interfaces, so removing the trajectory primarily harms quality rather than
savings.
Thus, across both model families, the HC trajectory is important for downshift quality: preserving it retains substantially more of HC's advantage than removing it entirely.

\paragraph{Takeaway.}
Downshift offers a useful operating point: LC models retain
a substantial share of HC's quality advantage while remaining cheaper than
HC-only. Removing the HC trajectory sharply reduces that retained
quality. Together with escalation, this reveals a directional duality: dropping
the sender's trajectory helps an HC receiver but harms an LC receiver.

\subsection{Cost Mechanics by Handoff Direction}
\label{sec:mechanics}

\begin{figure}[t]
\centering
\includegraphics[width=\linewidth]{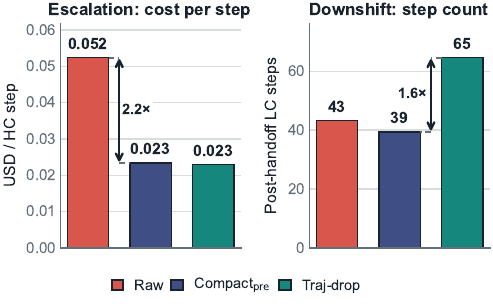}
\caption{\textbf{The handoff tax arises through different computational mechanisms.}
In Claude escalation, Raw makes each post-handoff HC step $2.2\times$ costlier
than Compact$_{\mathrm{pre}}$; in downshift, Traj-drop makes LC take
$1.6\times$ more steps. Full LC context inflates HC calls,
whereas missing HC context forces LC rework.}
\label{fig:fig_cost_decomp}
\end{figure}

Having established that trajectory removal improves escalation quality but
harms downshift quality across both families, we next examine how handoff
direction changes receiver-side computation and cost. We decompose post-handoff
cost into receiver step count and cost per step.
The Claude and GPT decompositions are shown in \Cref{fig:fig_cost_decomp} and Appendix~\ref{app:mechanics} (\Cref{fig:fig_cost_decomp_gpt}), respectively.

\paragraph{Raw escalation inflates the cost of each HC step.}
Relative to Compact$_{\mathrm{pre}}$, Raw raises the average post-handoff cost per HC step by $2.2\times$ for Claude and $1.6\times$ for GPT, while the two interfaces require a similar number of HC steps within each family. The
Raw-to-Compact$_{\mathrm{pre}}$ escalation premium therefore arises primarily
from more expensive receiver calls, not from a longer HC continuation.

\paragraph{Traj-drop downshift takes additional LC steps.}
Relative to Compact$_{\mathrm{pre}}$, Traj-drop downshift requires $1.6\times$ the post-handoff LC steps for Claude and $2.0\times$ for GPT,
while the two interfaces have similar costs per LC step within each
family. The
Traj-drop-to-Compact$_{\mathrm{pre}}$ downshift premium therefore arises
primarily from additional LC work, not from more expensive calls.
This is consistent with reconstructing missing HC context.

\paragraph{Takeaway.}
Raw escalation provides limited rescue, and the handoff tax depends on
direction and interface. Removing inherited LC trajectory improves quality for
HC receivers, whereas preserving inherited HC trajectory benefits LC receivers.
The associated cost penalties manifest through different computational channels:
more expensive HC steps under Raw escalation and additional LC steps under
Traj-drop downshift.

\section{Beyond the Coding-Agent Setting}
\label{sec:information-regimes}

\begin{figure}[t]
\centering
\includegraphics[width=\linewidth]{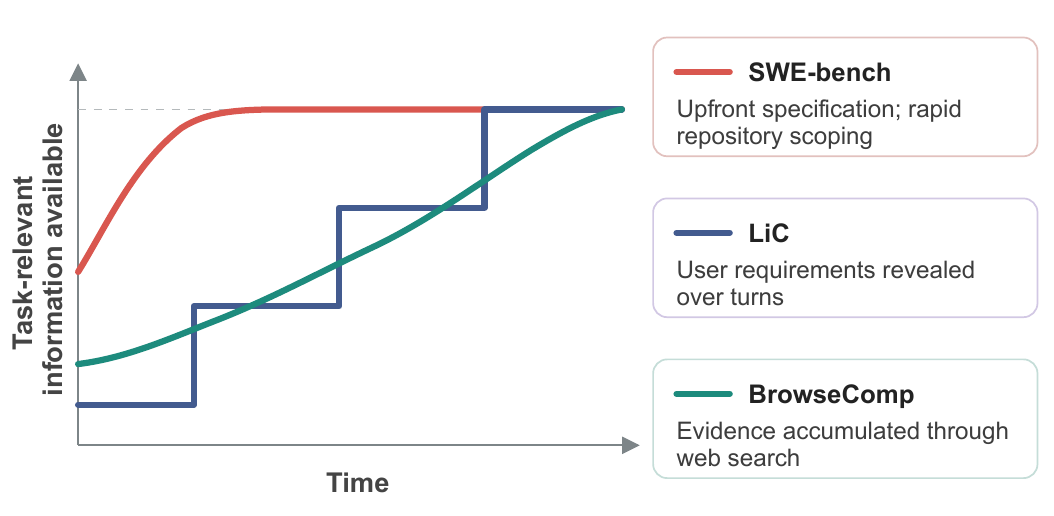}
\caption{\textbf{Task-relevant information evolves differently across settings.}
SWE-bench provides the specification upfront, followed by rapid repository
scoping; LiC reveals user requirements over turns; BrowseComp accumulates
evidence through web search. The schematic illustrates what information is
available at different stages of execution.}
\label{fig:task_information}
\end{figure}

Our primary study focuses on coding agents, one of the most consequential real-world settings for long-running agents, using SWE-bench, where the specification is available upfront and execution leaves persistent repository state.
We next vary an orthogonal property: the task's \emph{information dynamics}---how task-relevant information becomes available over time---which shape what the sender can accomplish before the switch and what remains for the receiver.
As illustrated in \Cref{fig:task_information}, SWE-bench provides the specification upfront while repository information accumulates during execution; LiC~\cite{laban2025llmslost} reveals
requirements incrementally across turns; and BrowseComp~\cite{browsecomp} provides the question upfront while evidence accumulates through search.
We evaluate Raw handoffs throughout to isolate this axis from interface design.

\begin{table}[t]
\centering
\small
\setlength{\tabcolsep}{2.4pt}
\renewcommand{\arraystretch}{1.06}
\resizebox{\columnwidth}{!}{%
\begin{tabular}{@{}llrrrr@{}}
  \toprule
  Tasks & Policy & Score & Cost & QRec & CSRet \\
  \midrule

  \multirow{4}{*}{
    \shortstack{
      \raisebox{-0.15em}{\includegraphics[height=0.85em]{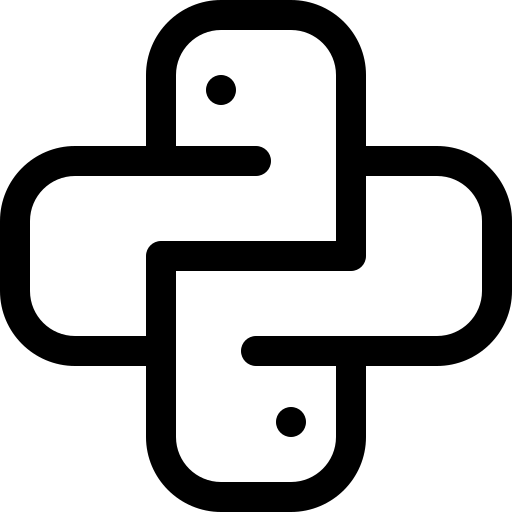}}\,
      \raisebox{-0.15em}{\includegraphics[height=0.85em]{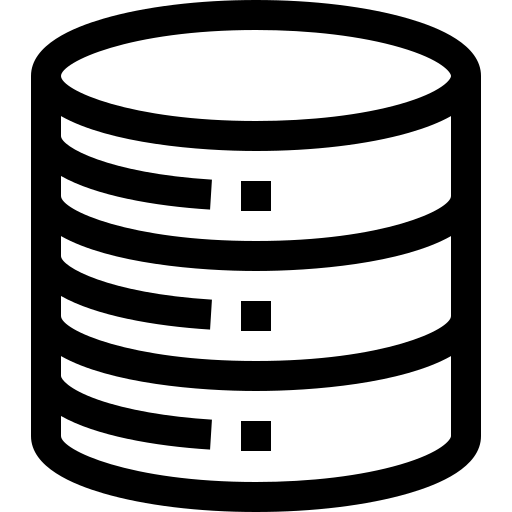}}\,
      \raisebox{-0.15em}{\includegraphics[height=0.85em]{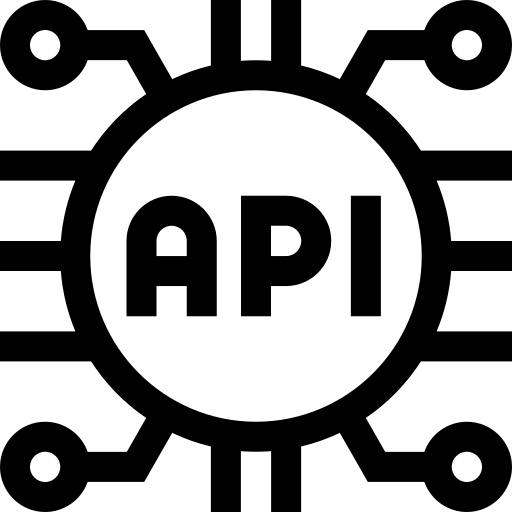}}
      \\[-0.1em]
      \raisebox{-0.15em}{\includegraphics[height=0.85em]{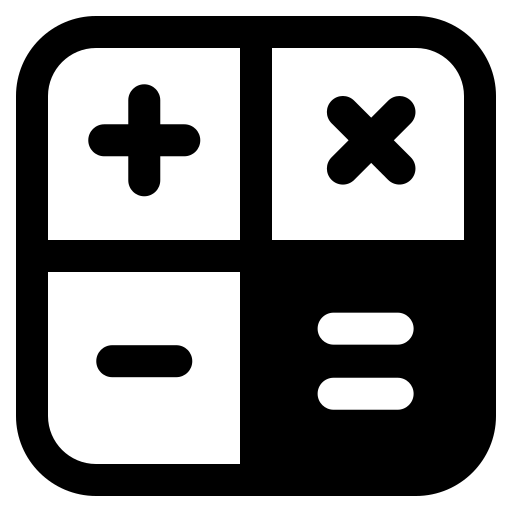}}\,
      \raisebox{-0.15em}{\includegraphics[height=0.85em]{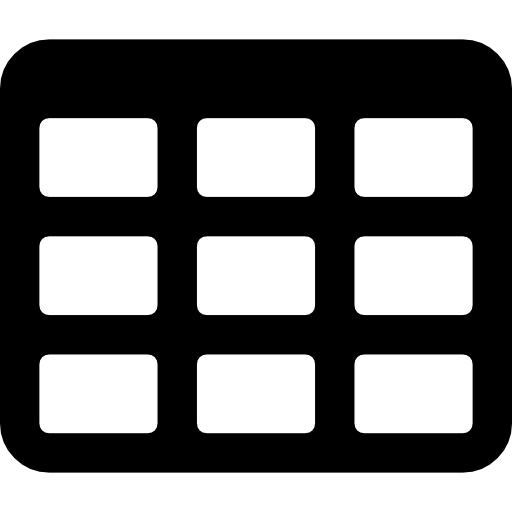}}
    }
  }
    & \cellcolor{gray!12}LC-only
    & \cellcolor{gray!12}62.1
    & \cellcolor{gray!12}0.015
    & \cellcolor{gray!12}$0$
    & \cellcolor{gray!12}$100$ \\

    & \cellcolor{gray!12}HC-only
    & \cellcolor{gray!12}76.9
    & \cellcolor{gray!12}0.080
    & \cellcolor{gray!12}$100$
    & \cellcolor{gray!12}$0$ \\

    & Escalation
    & 74.2
    & 0.056
    & $86$
    & $36$ \\

    & Downshift
    & 67.0
    & 0.046
    & $31$
    & $53$ \\

  \bottomrule
\end{tabular}%
}
\caption{\textbf{Late-arriving requirements favor an HC suffix.}
Raw-handoff results on LiC, averaged across five task families and early,
middle, and late structural switch positions for Claude.}
\label{tab:multiturn_raw_handoff}
\end{table}

\paragraph{Late-arriving requirements favor a strong suffix.}

We evaluate Raw handoffs on 535 LiC examples across five task families, where requirement shards progressively complete an initial high-level intent.
\Cref{tab:multiturn_raw_handoff} reports Claude averages across early, middle,
and late handoffs; GPT and position-specific results, together with the full
protocol, appear in \Cref{app:lic}.


Across all five families, late-arriving requirements reverse the coding quality
ordering while preserving the cost ordering: escalation recovers 86\% of HC's
quality advantage and retains 36\% of LC's cost advantage, versus 31\% and
53\% for downshift.
By construction, LiC remains underspecified until the final requirement shard,
so the receiver is the first model able to solve the fully specified task.
This favors escalation, which assigns that decisive stage to HC.



\begin{table}[t]
\centering
\small
\renewcommand{\arraystretch}{1.08}
\begin{tabular}{@{}lrrrr@{}}
\toprule
Strategy & QRec & CSRet & \multicolumn{2}{c}{Steps} \\
         &      &       & \multicolumn{1}{c}{LC}
                          & \multicolumn{1}{c}{HC} \\
\midrule
\textit{Full LC + full HC} & 100.0 & $-40.1$ & 62.6 & 33.4 \\
\textit{Abort + fresh HC}  & 100.0 &  $-8.5$ & 23.0 & 33.4 \\
\noalign{\vskip 2pt}
\hdashline
\noalign{\vskip 2pt}
Escalation                 &  95.8 & $-30.0$ & 23.0 & 30.4 \\
\midrule
Downshift                  &  56.7 &   76.8  & 22.8 & 17.0 \\
\bottomrule
\end{tabular}

\caption{\textbf{Under progressive search, escalation recovers quality but not savings.}
GPT Raw-handoff results on BrowseComp, averaged across switch points.}
\label{tab:browsecomp_raw_handoff}
\end{table}

\paragraph{Progressive search recovers quality but not savings.}

We evaluate Raw handoffs on 200 BrowseComp questions filtered to require web
browsing; the question is available upfront, but answer-relevant evidence
accumulates through search. \Cref{tab:browsecomp_raw_handoff} reports GPT
results. Full details in \Cref{app:browsecomp}.


For GPT, escalation nearly closes the HC quality gap ($\text{QRec}=95.8\%$), unlike
in SWE-bench, yet---as in SWE-bench---fails to produce savings: CSRet is
$-30.0\%$, worse than aborting LC and restarting HC from scratch ($-8.5\%$).
Downshift, by contrast, offers a useful middle ground, recovering 56.7\% of
HC's quality advantage while retaining 76.8\% of LC's cost advantage.
Escalation and \emph{Abort + fresh HC} share the same 23-step LC prefix, but
inheriting it shortens the HC continuation by three calls on average
(30.4 vs.\ 33.4), suggesting that earlier search progress remains useful.
Yet escalation remains substantially more expensive overall.

\paragraph{Takeaway.}

Together, these extensions show that handoff value depends on information
dynamics. LiC favors an HC receiver because the task becomes solvable only
after late requirements arrive; BrowseComp shows that inherited search progress
can nearly recover HC quality without producing savings. Thus, whether
escalation is worthwhile depends not only on model capability, but also on the
task-relevant state available at handoff.

\section{Discussion and Conclusion}
\label{sec:discussion}
Our results show that, under Raw handoff, escalation offers a poor cost--quality
trade-off, whereas downshift yields a favorable intermediate cost--quality point. This directional asymmetry extends to interface design: reducing
LC trajectory information improves escalation, whereas removing
HC trajectory information harms downshift. Together, these
suggest that model handoffs should be treated as a distinct
inference problem rather than merely as an extension of model routing. Routing determines which model acts next, whereas the handoff
interface determines what trajectory information that model inherits.
In long-horizon coding agents, a handoff
involves more than transferring conversational context: the receiver also
inherits a repository state produced by another model. It must continue both
an existing work product and a non-native trajectory, and how the trajectory
is transferred alongside that persistent state can substantially affect both
quality and cost. Thus, handoff design should be optimized
jointly with receiver selection and switch timing.

\paragraph{Future directions.}
We use fixed switch points to isolate the consequences of a handoff from the
policy that triggers it; future work can study adaptive, progress-aware
policies that jointly decide when to switch and which model should continue. Similarly, we evaluate a small set of
controlled interfaces to separate full trajectory transfer, compaction, and
trajectory removal; future work can develop structured handoffs that
selectively preserve, summarize, or discard particular trajectory components.
More broadly, routing methods should treat the handoff interface as part of
the switching policy rather than assume that the full trajectory simply
carries forward. Evaluation should also extend across model families,
capability gaps, pricing regimes, repeated rollouts, and multiple handoffs per trajectory.


\section*{Limitations}
We study two model pairs (Claude Haiku~4.5 / Opus~4.7 and GPT-5.6 Luna / Sol),
and our primary coding-agent study uses one benchmark, SWE-bench Verified.
Within this setting, however, we evaluate at substantial scale: 58
configurations per model family across all 500 tasks, totaling 58{,}000 agent
runs, 2 million LLM API calls, and 36 billion processed tokens.
To extend the scope of our study beyond the SWE-bench coding-agent setting, we evaluate a broader range of information dynamics but only under Raw transfer, so our interface findings are established primarily in the coding setting.
Switch points are fixed in advance using model- and
difficulty-calibrated percentiles of single-model trajectory length. The
resulting matched switched subsets are relatively small for the hard SWE-bench
stratum ($\bar{N}\!\approx\!24$--$27$ per reported cell), so we treat the
difficulty-conditioned findings as exploratory.
In the SWE-bench study, we run a single episode for each of the 500 tasks
under every configuration and therefore do not estimate variability across
repeated runs of the same task under the same configuration. To preserve
comparability despite this limitation, we restrict each interface comparison
to the shared intersection of tasks that switched under every strategy. We
also report task-clustered bootstrap confidence intervals, preserving each
task's observations across switch points. Cost comparisons, difficulty
analyses, and mechanism interpretations remain descriptive. Finally,
dollar-cost conclusions depend on provider pricing and prompt-cache rates.

\bibliography{custom}

\clearpage
\appendix
\section{SWE-bench Verified}
\label{sec:appendix}

\subsection{Experimental Details}

We provide the full SWE-bench Verified protocol, including the
agent environment, handoff implementation and prompts, switch-point
calibration, cost accounting, and software configuration.
\paragraph{Agent and task environment.}
We implement handoff by extending mini-SWE-agent with a wrapper containing exactly two models:
a prefix model \(M_{\mathrm{pre}}\) and a suffix model \(M_{\mathrm{suf}}\).
A single agent loop and Docker environment are retained throughout the run.
Consequently, the repository state, including all edits made by
\(M_{\mathrm{pre}}\), persists after handoff; the interfaces differ only in the
trajectory information supplied to \(M_{\mathrm{suf}}\). All models receive the same
SWE-bench task prompt and the same Bash tool, and command outputs are returned
to the model as tool observations. We evaluate all 500 test instances of
SWE-bench Verified.

\paragraph{Models and handoff directions.}
For Claude, the low-cost (LC) and high-capability (HC) models are Claude
Haiku~4.5 and Claude Opus~4.7, respectively. For GPT, they are GPT-5.6 Luna
and GPT-5.6 Sol. We study both escalation,
\(M_{\mathrm{LC}}\!\rightarrow M_{\mathrm{HC}}\), and downshift,
\(M_{\mathrm{HC}}\!\rightarrow M_{\mathrm{LC}}\).
Claude models were accessed through LiteLLM’s chat-completions interface, whereas GPT models were accessed through Bedrock’s OpenAI-compatible Responses endpoint; both used the same Bash tool, with conversation histories translated into each endpoint’s native tool-call format.
For sampling, we set temperature to 0 for Haiku; for GPT-5.6 and Opus 4.7, we use provider defaults, with \emph{medium} and \emph{high} reasoning effort, respectively.

\paragraph{Switch rule.}
Let \(q=0,1,\ldots\) index agent model calls and let \(s_b\) be the switch point
for difficulty bucket \(b\). Calls \(q<s_b\) use \(M_{\mathrm{pre}}\), and call
\(q=s_b\) is the first call to \(M_{\mathrm{suf}}\). Thus, \(s_b\) prefix
actions and their resulting observations are complete before the suffix model
is queried. Thresholds are fixed before the handoff runs at percentiles
\(p\in\{5,10,15,25,35,45,50\}\) of the prefix model's single-model
termination-step distribution. They are computed separately for easy
(\(<15\) minute), medium (\(15\) minutes--\(1\) hour), and hard
(\(>1\) hour) tasks; the two longest SWE-bench difficulty categories are
merged into the hard bucket.
\Cref{tab:step-distribution} report the exact
thresholds.

\paragraph{Handoff interfaces.}
Let $\mathcal{T}_{1:K}$ denote the trajectory produced by the prefix model up
to step $K$, and let $\mathcal{W}_K$ denote the persistent working tree at
that point. All interfaces preserve $\mathcal{W}_K$; they differ only in the
trajectory information passed to the suffix model. We compare four principal
interfaces:
\begin{description}
  \item[Raw.] The suffix model receives the full trajectory
  $\mathcal{T}_{1:K}$ together with $\mathcal{W}_K$: the system prompt, task,
  prefix reasoning, tool calls, and tool observations are retained verbatim.
  No message announces the model change.

  \item[Compact$_{\mathrm{pre}}$.] The prefix model reads
  $\mathcal{T}_{1:K}$ and writes a plain-text continuation summary. The suffix
  model receives only this summary together with $\mathcal{W}_K$.

  \item[Compact$_{\mathrm{suf}}$.] The suffix model reads
  $\mathcal{T}_{1:K}$ and writes the continuation summary itself, allowing us
  to isolate which model should author the handoff representation. It then
  continues from this summary together with $\mathcal{W}_K$.

  \item[Traj-drop.] No trajectory information from
  $\mathcal{T}_{1:K}$ is transferred, and no summarization call is made. The
  suffix model receives only the system prompt, original task, a static
  continuation message, and $\mathcal{W}_K$.
\end{description}

For both compact interfaces, the summarizer is given the full prefix
trajectory followed by:

\begin{promptbox}{Compaction instruction}
{\fontsize{8.5pt}{10pt}\selectfont\ttfamily\RaggedRight
You are summarizing a coding agent trajectory for handing off work to another
coding agent.\par\medskip
The conversation above is a coding agent's work-in-progress trajectory on a
task.\par\medskip
Write a summary of what the agent did so a different agent---with no access to
this conversation---can continue the task.
\par}
\end{promptbox}

Let $\widehat{\mathcal{T}}_{1:K}$ denote the resulting summary. We then replace
the live context with
\[
  [\,m_{\mathrm{system}},\,m_{\mathrm{task}},\,
    m_{\mathrm{handoff}}(\widehat{\mathcal{T}}_{1:K})\,],
\]
where $m_{\mathrm{handoff}}$ uses the following message:

\begin{promptbox}{Compaction continuation message}
{\fontsize{8.5pt}{10pt}\selectfont\ttfamily\RaggedRight
Below is a summary of a previous agent's work on this task. Any file changes
it made are still in the working tree.\par\medskip
\textless\textless SUMMARY\textgreater\textgreater
\par}
\end{promptbox}

\noindent Here, \texttt{<<SUMMARY>>} is replaced by
$\widehat{\mathcal{T}}_{1:K}$. The suffix model then produces its first
executable action.
Traj-drop uses the same three-message structure but replaces the summary with
the following fixed continuation message:

\begin{promptbox}{Traj-drop continuation message}
{\fontsize{8.5pt}{10pt}\selectfont\ttfamily\RaggedRight
A previous agent worked on this task. Any file changes it made are still in
the working tree.\par\medskip
Continue solving the task.\par}
\end{promptbox}

\noindent The removed trajectory is written to an audit artifact but is never exposed
to the suffix model. 
The restart controls are composed from existing matched runs rather than
executed as separate trajectories. \emph{Abort + HC fresh} combines the
observed LC prefix cost and steps through $K$ with the outcome, cost, and steps
of the matched HC-only run. \emph{LC-full + HC-full} similarly combines the
cost and steps of the complete LC-only and HC-only runs. In both controls,
quality is given by the matched HC-only outcome.


\paragraph{Accounting.}
Each response is tagged with its active model, zero-indexed call number,
prefix/suffix phase, effective threshold, and task difficulty. The compact interfaces make one paid summarization call; its cost and usage
are attributed to the handoff and folded into the first suffix step, although
it is not counted as an additional environment-action step. Traj-drop and Raw
introduce no additional model calls. Costs are reconstructed from
provider-reported input, cache, and output-token usage using the corresponding
cache-aware price cards. Output-token counts include reasoning tokens.
Billable malformed responses are retained and counted. Runs use a 150-step
cap and are audited so that prefix, handoff, and suffix costs reconcile exactly
with total instance cost. Final patches are scored with the official
SWE-bench evaluator.

\paragraph{Pricing and audit.}
Token prices follow LiteLLM v1.83.14's model-pricing registry for the
configured Bedrock model identifiers, including cache pricing and
cross-region premiums where applicable. The GPT-5.6 Bedrock identifiers use a
pinned LiteLLM-compatible local price card distributed with our experiment
configuration. The accompanying artifact records the exact invocation
identifiers, price-card entries, and implementation and evaluator revisions.
Before reporting results, we verify that every required task--condition run
has an audited terminal trajectory and evaluator outcome. Infrastructure
failures are rerun rather than scored, and no task or condition is excluded.
Aggregate retry-attempt counts were not retained.

\subsection{Switch-Point Calibration}
\label{app:swe_switch_calibration}

\Cref{tab:step-distribution} reports the exact difficulty-calibrated switch
steps used in the SWE-bench experiments for both model families and handoff
directions. Each threshold is the corresponding percentile of the prefix
model's single-model termination-step distribution within a difficulty bucket.
Escalation therefore uses thresholds calibrated from LC trajectories, whereas
downshift uses thresholds calibrated from HC trajectories.

\begin{figure}[t]
\centering
\includegraphics[width=\linewidth]{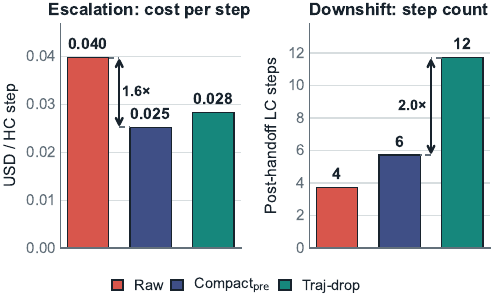}
\caption{\textbf{GPT exhibits the same handoff-cost accounting pattern as
Claude.}
Relative to Compact$_{\mathrm{pre}}$, each post-handoff HC step under Raw costs
$1.6\times$ as much, while Traj-drop downshift takes $2.0\times$ as many
post-handoff LC steps.
}
\label{fig:fig_cost_decomp_gpt}
\end{figure}

\subsection{Additional Results and Analyses}
\label{app:breakdowns}
\label{app:breakdowns_swe}

\paragraph{Full results across model families.}
\Cref{tab:handoff_duality} reports the complete aggregate results for the
Claude and GPT model pairs on matched switched subsets. Across both families,
Raw escalation recovers less than half of the LC--HC quality gap, whereas
downshift occupies a favorable intermediate cost--quality point. Reducing
LC trajectory information improves escalation quality, while
preserving HC trajectory information improves downshift quality.
The analyses below disaggregate and test these aggregate patterns.

\paragraph{Difficulty breakdown.}
\Cref{tab:difficulty_handoff} conditions the Claude results on the SWE-bench
Verified difficulty buckets, providing the numbers behind
\Cref{fig:practitioner-difficulty}. Escalation economics improve with
difficulty, with the structured interfaces recovering more quality at a more
favorable cost. In downshift, the context-preserving interfaces consistently
outperform Traj-drop in quality recovery and cost efficiency. Results for the
small hard-task subset should be interpreted directionally.

\begin{table*}
\centering
\small
\begin{minipage}[t]{0.49\textwidth}
\centering
\textbf{(a) Escalation}\\[-0.1em]
\setlength{\tabcolsep}{3pt}
\renewcommand{\arraystretch}{1.08}
\resizebox{\linewidth}{!}{%
  \begin{tabular}{llrrrrrr}
    \toprule
Strategy & Difficulty
  & $\bar{N}$ & Pass
  & $\Delta_{\mathrm{LC}}$
  & $\Delta_{\mathrm{HC}}$
  & QRec
  & CSRet \\
& & & (\%) & (pp) & (pp) & (\%) & (\%) \\
    \midrule
    Raw & Easy & 129 & 82.1 & +3.9 & -6.0 & 38 & -982 \\
     & Medium & 160 & 63.6 & +10.8 & -11.7 & 49 & -275 \\
     & Hard & 24 & 32.5 & +19.2 & -21.4 & 47 & -86 \\
    \midrule
    Compact$_{\text{pre}}$ & Easy & 129 & 83.4 & \textbf{+5.2} & -4.7 & \textbf{53} & \textbf{-164} \\
     & Medium & 160 & 66.1 & +13.3 & -9.2 & 59 & \textbf{-12} \\
     & Hard & 24 & 42.7 & \textbf{+29.5} & -11.2 & \textbf{74} & \textbf{42} \\
    \midrule
    Compact$_{\text{suf}}$ & Easy & 129 & 82.2 & +4.1 & -5.8 & 41 & -482 \\
     & Medium & 160 & 63.1 & +10.3 & -12.2 & 47 & -68 \\
     & Hard & 24 & 39.6 & +26.4 & -14.3 & 65 & 12 \\
     \midrule
    Traj-drop & Easy & 129 & 81.9 & +3.7 & -6.2 & 38 & -213 \\
     & Medium & 160 & 68.6 & \textbf{+15.9} & -6.7 & \textbf{71} & -31 \\
     & Hard & 24 & 42.3 & +29.0 & -11.6 & 72 & 34 \\
    \bottomrule
  \end{tabular}
}
\end{minipage}
\hfill
\begin{minipage}[t]{0.49\textwidth}
\centering
\textbf{(b) Downshift}\\[-0.1em]
\setlength{\tabcolsep}{3pt}
\renewcommand{\arraystretch}{1.08}
\resizebox{\linewidth}{!}{%
  \begin{tabular}{llrrrrrr}
    \toprule
Strategy & Difficulty
  & $\bar{N}$ & Pass
  & $\Delta_{\mathrm{LC}}$
  & $\Delta_{\mathrm{HC}}$
  & QRec
  & CSRet \\
& & & (\%) & (pp) & (pp) & (\%) & (\%) \\
    \midrule
    Raw & Easy & 113 & 78.0 & +3.0 & -7.6 & 23 & \textbf{46} \\
     & Medium & 157 & 61.8 & +15.2 & -9.8 & 59 & \textbf{81} \\
     & Hard & 27 & 35.6 & +20.6 & -22.9 & 46 & \textbf{88} \\
    \midrule
    Compact$_{\text{pre}}$ & Easy & 113 & 79.0 & \textbf{+4.0} & -6.6 & \textbf{35} & 44 \\
     & Medium & 157 & 62.5 & \textbf{+15.8} & -9.2 & \textbf{62} & 80 \\
     & Hard & 27 & 40.4 & \textbf{+25.4} & -18.1 & \textbf{57} & 87 \\
    \midrule
    Compact$_{\text{suf}}$ & Easy & 113 & 76.2 & +1.2 & -9.4 & 6 & 28 \\
     & Medium & 157 & 59.1 & +12.4 & -12.5 & 48 & 74 \\
     & Hard & 27 & 38.8 & +23.7 & -19.8 & 56 & 86 \\
     \midrule
    Traj-drop & Easy & 113 & 72.8 & -2.2 & -12.8 & -28 & -31 \\
     & Medium & 157 & 57.4 & +10.7 & -14.2 & 42 & 59 \\
     & Hard & 27 & 31.3 & +16.2 & -27.2 & 38 & 82 \\
    \bottomrule
  \end{tabular}
}
\end{minipage}

\caption{\textbf{Difficulty-conditioned handoff results.}
Claude intersection-only results averaged uniformly over switch points p5--p50, split by
SWE-bench Verified difficulty. $\Delta_{\mathrm{LC}}$ is the pass-rate gain over
LC-only; $\Delta_{\mathrm{HC}}$ is the pass-rate difference from HC-only, with
negative values indicating remaining gap. QRec and CSRet follow
Eqs.~\ref{eq:qrec}--\ref{eq:csret}. Bold marks the best
$\Delta_{\mathrm{LC}}$, QRec, and CSRet within each difficulty bucket and direction.}
\label{tab:difficulty_handoff}
\end{table*}

\paragraph{Early and late switch breakdown.}
\Cref{tab:handoff-timing-esc,tab:handoff-timing-dn} split the
Claude and GPT results into Early (p5--p15) and Late (p25--p50)
switch windows. The central interface reversal is stable across both
timing windows and model families: Traj-drop achieves the highest
quality recovery in escalation and the lowest in downshift. Switch
timing changes the magnitude, but not the direction, of this effect.
Later escalation generally recovers less quality, whereas later
downshift improves quality recovery for Raw and both compaction
interfaces while retaining less savings. The finer ordering among
the context-preserving interfaces varies, particularly for GPT.

\begin{table*}
\centering
\small
\setlength{\tabcolsep}{2.6pt}
\renewcommand{\arraystretch}{1.0}
\resizebox{\textwidth}{!}{%
  \begin{tabular}{@{}lrrrrrlrrrrr@{}}
    \toprule
    \multicolumn{6}{c}{\textbf{Claude} (Haiku~4.5 / Opus~4.7)} &
    \multicolumn{6}{c}{\textbf{GPT-5.6} (Luna / Sol)} \\
    \cmidrule(lr){1-6}\cmidrule(lr){7-12}
    Strategy & Pass & Cost & Steps & QRec & CSRet &
    \hspace{9pt}Strategy & Pass & Cost & Steps & QRec & CSRet \\
    \midrule
    \multicolumn{12}{l}{\textit{All}} \\
    \rowcolor{gray!12} LC-only & 60.7 & 0.40 & 80 & $0$ & $100$ & \hspace{9pt}LC-only & 58.7 & 0.06 & 14 & $0$ & $100$ \\
    \rowcolor{gray!12} HC-only & 79.2 & 0.72 & 31 & $100$ & $0$ & \hspace{9pt}HC-only & 83.7 & 0.47 & 17 & $100$ & $0$ \\
    \noalign{\vskip 2pt}
    \cdashline{1-12}
    \noalign{\vskip 2pt}
    \rowcolor{gray!12} Abort + HC fresh & 79.2$^\dagger$ & 0.90 & 78 & $100^\dagger$ & $-58$ & \hspace{9pt}Abort + HC fresh & 83.7$^\dagger$ & 0.51 & 27 & $100^\dagger$ & $-11$ \\
    \rowcolor{gray!12} LC-full + HC-full & 79.2$^\dagger$ & 1.12 & 111 & $100^\dagger$ & $-130$ & \hspace{9pt}LC-full + HC-full & 83.7$^\dagger$ & 0.53 & 31 & $100^\dagger$ & $-14$ \\
    \midrule
    Raw handoff & 69.2 & 1.61 & 74 & $47$ & $-285$ & \hspace{9pt}Raw handoff & 67.5 & 0.36 & \textbf{17} & $36$ & $27$ \\
    Compact$_{\text{pre}}$ & 71.8 & \textbf{0.75} & 71 & $60$ & $\mathbf{-11}$ & \hspace{9pt}Compact$_{\text{pre}}$ & 68.8 & \textbf{0.27} & 18 & $40$ & $\mathbf{49}$ \\
    Compact$_{\text{suf}}$ & 69.6 & 0.98 & \textbf{67} & $49$ & $-82$ & \hspace{9pt}Compact$_{\text{suf}}$ & 68.8 & 0.43 & 19 & $40$ & $10$ \\
    Traj-drop & \textbf{72.4} & 0.81 & 74 & $\mathbf{64}$ & $-30$ & \hspace{9pt}Traj-drop & \textbf{79.7} & 0.50 & 25 & $\mathbf{84}$ & $-8$ \\
    \specialrule{\lightrulewidth}{3pt}{3pt}
    \multicolumn{12}{l}{\textit{Early}} \\
    \rowcolor{gray!12} LC-only & 63.5 & 0.36 & 73 & $0$ & $100$ & \hspace{9pt}LC-only & 63.2 & 0.05 & 13 & $0$ & $100$ \\
    \rowcolor{gray!12} HC-only & 80.4 & 0.64 & 29 & $100$ & $0$ & \hspace{9pt}HC-only & 83.7 & 0.42 & 17 & $100$ & $0$ \\
    \noalign{\vskip 2pt}
    \cdashline{1-12}
    \noalign{\vskip 2pt}
    \rowcolor{gray!12} Abort + HC fresh & 80.4$^\dagger$ & 0.78 & 65 & $100^\dagger$ & $-49$ & \hspace{9pt}Abort + HC fresh & 83.7$^\dagger$ & 0.46 & 25 & $100^\dagger$ & $-10$ \\
    \rowcolor{gray!12} LC-full + HC-full & 80.4$^\dagger$ & 1.00 & 102 & $100^\dagger$ & $-129$ & \hspace{9pt}LC-full + HC-full & 83.7$^\dagger$ & 0.47 & 29 & $100^\dagger$ & $-13$ \\
    \midrule
    Raw handoff & 73.1 & 1.38 & 64 & $57$ & $-265$ & \hspace{9pt}Raw handoff & 72.3 & 0.32 & \textbf{16} & $44$ & $27$ \\
    Compact$_{\text{pre}}$ & 74.7 & \textbf{0.66} & 60 & $66$ & $\mathbf{-6}$ & \hspace{9pt}Compact$_{\text{pre}}$ & 71.7 & \textbf{0.24} & 16 & $42$ & $\mathbf{49}$ \\
    Compact$_{\text{suf}}$ & 73.6 & 0.82 & \textbf{56} & $60$ & $-65$ & \hspace{9pt}Compact$_{\text{suf}}$ & 71.4 & 0.37 & 17 & $40$ & $15$ \\
    Traj-drop & \textbf{75.0} & 0.73 & 62 & $\mathbf{69}$ & $-32$ & \hspace{9pt}Traj-drop & \textbf{81.2} & 0.45 & 23 & $\mathbf{87}$ & $-8$ \\
    \specialrule{\lightrulewidth}{3pt}{3pt}
    \multicolumn{12}{l}{\textit{Late}} \\
    \rowcolor{gray!12} LC-only & 58.6 & 0.44 & 84 & $0$ & $100$ & \hspace{9pt}LC-only & 55.4 & 0.07 & 15 & $0$ & $100$ \\
    \rowcolor{gray!12} HC-only & 78.3 & 0.77 & 34 & $100$ & $0$ & \hspace{9pt}HC-only & 83.8 & 0.50 & 18 & $100$ & $0$ \\
    \noalign{\vskip 2pt}
    \cdashline{1-12}
    \noalign{\vskip 2pt}
    \rowcolor{gray!12} Abort + HC fresh & 78.3$^\dagger$ & 0.99 & 87 & $100^\dagger$ & $-66$ & \hspace{9pt}Abort + HC fresh & 83.8$^\dagger$ & 0.55 & 28 & $100^\dagger$ & $-12$ \\
    \rowcolor{gray!12} LC-full + HC-full & 78.3$^\dagger$ & 1.21 & 118 & $100^\dagger$ & $-131$ & \hspace{9pt}LC-full + HC-full & 83.8$^\dagger$ & 0.56 & 32 & $100^\dagger$ & $-15$ \\
    \midrule
    Raw handoff & 66.4 & 1.78 & 82 & $39$ & $-299$ & \hspace{9pt}Raw handoff & 63.9 & 0.39 & \textbf{18} & $30$ & $26$ \\
    Compact$_{\text{pre}}$ & 69.6 & \textbf{0.82} & 79 & $56$ & $\mathbf{-16}$ & \hspace{9pt}Compact$_{\text{pre}}$ & 66.6 & \textbf{0.29} & 18 & $39$ & $\mathbf{49}$ \\
    Compact$_{\text{suf}}$ & 66.6 & 1.09 & \textbf{74} & $40$ & $-95$ & \hspace{9pt}Compact$_{\text{suf}}$ & 66.9 & 0.47 & 20 & $41$ & $7$ \\
    Traj-drop & \textbf{70.4} & 0.87 & 82 & $\mathbf{60}$ & $-28$ & \hspace{9pt}Traj-drop & \textbf{78.6} & 0.53 & 27 & $\mathbf{82}$ & $-8$ \\
    \bottomrule
  \end{tabular}
}
\caption{\textbf{Coding-agent escalation by switch-timing window.}
Columns compare model families. \emph{All} averages p5--p50, \emph{Early}
averages \{p5,p10,p15\}, and \emph{Late} averages
\{p25,p35,p45,p50\}. Values are unweighted means over switch points on
matched switched subsets. Pass is in percent, Cost in dollars, and Steps is
the mean trajectory length. QRec is recovered LC--HC quality gap; CSRet is
retained cost savings, with negative values costlier than HC-only. Gray rows
are single-model baselines and restart references. Bold marks the best
available handoff quality and best non-baseline cost/efficiency within each
family and timing window.}
\label{tab:handoff-timing}
\label{tab:handoff-timing-esc}
\end{table*}

\begin{table*}
\centering
\small
\setlength{\tabcolsep}{2.6pt}
\renewcommand{\arraystretch}{1.0}
\resizebox{\textwidth}{!}{%
  \begin{tabular}{@{}lrrrrrlrrrrr@{}}
    \toprule
    \multicolumn{6}{c}{\textbf{Claude} (Haiku~4.5 / Opus~4.7)} &
    \multicolumn{6}{c}{\textbf{GPT-5.6} (Luna / Sol)} \\
    \cmidrule(lr){1-6}\cmidrule(lr){7-12}
    Strategy & Pass & Cost & Steps & QRec & CSRet &
    \hspace{9pt}Strategy & Pass & Cost & Steps & QRec & CSRet \\
    \midrule
    \multicolumn{12}{l}{\textit{All}} \\
    \rowcolor{gray!12} LC-only & 54.6 & 0.41 & 79 & $0$ & $100$ & \hspace{9pt}LC-only & 63.6 & 0.05 & 13 & $0$ & $100$ \\
    \rowcolor{gray!12} HC-only & 75.8 & 0.85 & 35 & $100$ & $0$ & \hspace{9pt}HC-only & 85.8 & 0.47 & 18 & $100$ & $0$ \\
    \midrule
    Raw handoff & 65.6 & \textbf{0.51} & 58 & $50$ & $\mathbf{80}$ & \hspace{9pt}Raw handoff & \textbf{81.0} & \textbf{0.42} & \textbf{16} & $\mathbf{79}$ & $\mathbf{14}$ \\
    Compact$_{\text{pre}}$ & \textbf{66.8} & 0.52 & \textbf{54} & $\mathbf{56}$ & $78$ & \hspace{9pt}Compact$_{\text{pre}}$ & 79.8 & 0.43 & 18 & $72$ & $10$ \\
    Compact$_{\text{suf}}$ & 63.7 & 0.53 & 65 & $42$ & $73$ & \hspace{9pt}Compact$_{\text{suf}}$ & 80.4 & 0.42 & 17 & $75$ & $13$ \\
    Traj-drop & 60.9 & 0.59 & 79 & $28$ & $59$ & \hspace{9pt}Traj-drop & 75.4 & 0.43 & 24 & $53$ & $10$ \\
    \specialrule{\lightrulewidth}{3pt}{3pt}
    \multicolumn{12}{l}{\textit{Early}} \\
    \rowcolor{gray!12} LC-only & 62.5 & 0.36 & 72 & $0$ & $100$ & \hspace{9pt}LC-only & 66.5 & 0.05 & 12 & $0$ & $100$ \\
    \rowcolor{gray!12} HC-only & 80.3 & 0.66 & 30 & $100$ & $0$ & \hspace{9pt}HC-only & 86.0 & 0.41 & 17 & $100$ & $0$ \\
    \midrule
    Raw handoff & 69.0 & \textbf{0.41} & 55 & $35$ & $\mathbf{83}$ & \hspace{9pt}Raw handoff & \textbf{82.3} & 0.34 & \textbf{15} & $\mathbf{81}$ & $20$ \\
    Compact$_{\text{pre}}$ & \textbf{71.0} & \textbf{0.41} & \textbf{49} & $\mathbf{47}$ & $\mathbf{83}$ & \hspace{9pt}Compact$_{\text{pre}}$ & 79.9 & 0.36 & 17 & $69$ & $16$ \\
    Compact$_{\text{suf}}$ & 67.6 & 0.44 & 60 & $28$ & $75$ & \hspace{9pt}Compact$_{\text{suf}}$ & 79.9 & \textbf{0.34} & 16 & $69$ & $\mathbf{21}$ \\
    Traj-drop & 65.9 & 0.48 & 72 & $18$ & $59$ & \hspace{9pt}Traj-drop & 77.2 & 0.35 & 22 & $55$ & $18$ \\
    \specialrule{\lightrulewidth}{3pt}{3pt}
    \multicolumn{12}{l}{\textit{Late}} \\
    \rowcolor{gray!12} LC-only & 48.6 & 0.45 & 85 & $0$ & $100$ & \hspace{9pt}LC-only & 61.4 & 0.06 & 14 & $0$ & $100$ \\
    \rowcolor{gray!12} HC-only & 72.4 & 0.98 & 40 & $100$ & $0$ & \hspace{9pt}HC-only & 85.7 & 0.51 & 19 & $100$ & $0$ \\
    \midrule
    Raw handoff & 63.1 & \textbf{0.58} & 60 & $61$ & $\mathbf{77}$ & \hspace{9pt}Raw handoff & 79.9 & \textbf{0.47} & \textbf{17} & $77$ & $\mathbf{9}$ \\
    Compact$_{\text{pre}}$ & \textbf{63.6} & 0.59 & \textbf{58} & $\mathbf{63}$ & $74$ & \hspace{9pt}Compact$_{\text{pre}}$ & 79.8 & 0.49 & 19 & $75$ & $5$ \\
    Compact$_{\text{suf}}$ & 60.9 & 0.60 & 69 & $52$ & $71$ & \hspace{9pt}Compact$_{\text{suf}}$ & \textbf{80.7} & 0.48 & 18 & $\mathbf{80}$ & $8$ \\
    Traj-drop & 57.1 & 0.68 & 85 & $36$ & $59$ & \hspace{9pt}Traj-drop & 74.0 & 0.50 & 26 & $52$ & $4$ \\
    \bottomrule
  \end{tabular}
}
\caption{\textbf{Coding-agent downshift by switch-timing window.}
Columns compare model families. \emph{All} averages p5--p50, \emph{Early}
averages \{p5,p10,p15\}, and \emph{Late} averages
\{p25,p35,p45,p50\}. Values are unweighted means over switch points on
matched switched subsets. Pass is in percent, Cost in dollars, and Steps is
the mean trajectory length. QRec is recovered LC--HC quality gap; CSRet is
retained cost savings. Gray rows are single-model baselines. Bold marks the
best available handoff quality and best non-baseline cost/efficiency within
each family and timing window.}
\label{tab:handoff-timing-dn}
\end{table*}


\paragraph{Receiver-side computation.}
\label{app:mechanics}


Figure~\ref{fig:fig_cost_decomp_gpt} reports the GPT decomposition
corresponding to the Claude analysis in \Cref{sec:mechanics}.
The same pattern holds across model families.
In escalation, Raw increases the average cost of each post-handoff HC step
relative to Compact$_{\mathrm{pre}}$, while requiring a similar number of HC steps.
In downshift, Traj-drop requires substantially more post-handoff LC steps
than Compact$_{\mathrm{pre}}$, while the interfaces have similar costs per LC step.
Thus, the Raw escalation premium arises primarily from more expensive receiver
calls, whereas the Traj-drop downshift premium arises primarily from additional
LC work.

\paragraph{Model-change disclosure ablation.}
\label{app:disclosure}

In raw handoff, a natural question is whether simply \emph{telling} the receiver that the preceding work
was produced by a different model recovers part of the escalation tax. To
isolate the effect of model-change disclosure from context transformation, we
compare Raw escalation with a disclosure-only variant using Claude. Both
conditions preserve the full trajectory without summarization; the disclosure
condition appends a single user message immediately before the suffix model's
first call:

\begin{promptbox}{
  Handoff disclosure
  \hfill
  {\normalfont\footnotesize Injected user message}
}
{\fontsize{8.5pt}{10pt}\selectfont
\ttfamily
\RaggedRight
The preceding assistant messages contain a previous agent's work on this task.
The previous agent used a different model. Any file changes it made are still
in the working tree. Continue solving the original task.
\par}
\end{promptbox}

Across p5--p50, averaged uniformly over switch points on the shared fired
intersection---here recomputed to include the disclosure condition, so the
baselines differ slightly from \Cref{tab:handoff_duality}---disclosure provides
a modest quality gain at a slightly higher cost (\Cref{tab:raw_disclosure}).
It remains below HC-only and is dominated in both quality and cost by
Compact$_{\text{pre}}$ and Traj-drop. 
This disclosure-only prompt does not eliminate the Raw handoff tax in Claude escalation.

\begin{table}[H]
  \centering
  \setlength{\tabcolsep}{4pt}\renewcommand{\arraystretch}{1.15}
  \resizebox{\columnwidth}{!}{%
  \begin{tabular}{@{}l rrr rr@{}}
    \toprule
    Strategy & Pass & Cost & Steps & QRec & CSRet\\
             & (\%) & (\$) & (\#)  & (\%) & (\%)\\
    \midrule
    \cellcolor{gray!12}LC-only & \cellcolor{gray!12}60.6 & \cellcolor{gray!12}0.41 & \cellcolor{gray!12}80 & \cellcolor{gray!12}$0$ & \cellcolor{gray!12}$100$ \\
    \cellcolor{gray!12}HC-only & \cellcolor{gray!12}78.8 & \cellcolor{gray!12}0.72 & \cellcolor{gray!12}31 & \cellcolor{gray!12}$100$ & \cellcolor{gray!12}$0$ \\
    \cmidrule(l){1-6}
    Raw handoff & 69.3 & 1.62 & 74 & $48$ & $-285$ \\
    Raw + disclosure & 70.5 & 1.67 & 75 & $55$ & $-300$ \\
    \cdashline{1-6}
    Compact$_{\text{pre}}$ & 71.5 & \textbf{0.75} & 71 & $60$ & $\mathbf{-10}$ \\
    Compact$_{\text{suf}}$ & 69.6 & 0.98 & \textbf{67} & $50$ & $-83$ \\
    Traj-drop & \textbf{72.1} & 0.94 & 77 & $\mathbf{64}$ & $-72$ \\
    \bottomrule
  \end{tabular}}
  \caption{\textbf{Escalation (LC~$\rightarrow$~HC), averaged over all switch points.} Intersection-only results average p5--p50 unweighted. \textbf{QRec} is quality-gap recovery and \textbf{CSRet} is cost-savings retention; both are 100\% at the ideal corner. \textbf{LC}=Haiku~4.5 and \textbf{HC}=Opus~4.7. Compact$_{\text{pre}}$/Compact$_{\text{suf}}$ use the handoff-from/handoff-to model to summarize. Raw + disclosure retains the full trajectory and appends one model-change notice at the first HC call. Best Pass/QRec among handoff strategies and best Cost/Steps/CSRet among all non-baseline methods are in \textbf{bold}.}
  \label{tab:raw_disclosure}
\end{table}

\paragraph{Statistical uncertainty.}
We quantify task-to-task variation in pairwise interface pass-rate differences.
At each switch point, differences are computed on the shared subset of tasks
that reached the handoff under all four interfaces. The seven switch-point
differences are then averaged with equal weight, matching the aggregation in
the main table.

For each model family and direction, we collect the unique task IDs appearing
in at least one of the seven matched subsets. These pools contain 460
escalation and 458 downshift tasks for Claude, and 411 escalation and 426
downshift tasks for GPT. Each bootstrap replicate draws the same number of task
IDs with replacement. When a task is drawn more than once, it receives the
same weight under all interfaces and at every switch point where it appears.
We then recompute the seven pass-rate differences and their equal-weight
average. We report pointwise 95\% confidence intervals from $10{,}000$
bootstrap replicates.

\Cref{tab:sig-primary-combined} presents the central Traj-drop--Raw contrasts
first within each direction, followed by Traj-drop--compaction contrasts and
secondary comparisons among the context-preserving interfaces. The central
reversal is consistent across both families: Traj-drop improves escalation
pass rate and reduces downshift pass rate relative to Raw, with all four
pointwise intervals excluding zero. The broader Traj-drop--compaction
contrasts follow the same directional pattern, whereas comparisons among Raw
and the two compaction interfaces vary across families. In particular, we do
not treat summary authorship as a cross-family finding.

\begin{table*}
\centering
\small
\renewcommand{\arraystretch}{1.08}
\begin{tabular}{@{}lll rc rc@{}}
\toprule
\multirow{2}{*}{Direction}
& \multicolumn{2}{c}{Comparison}
& \multicolumn{2}{c}{Claude}
& \multicolumn{2}{c}{GPT} \\
\cmidrule(lr){2-3}\cmidrule(lr){4-5}\cmidrule(lr){6-7}
& First & Second
& $\Delta$Pass & 95\% CI
& $\Delta$Pass & 95\% CI \\
\midrule
\multirow{6}{*}{Escalation}
& Traj-drop & Raw
& $+3.1$ & $[+1.3,+5.0]$
& $+12.2$ & $[+9.2,+15.4]$ \\
\addlinespace[2pt]
& Traj-drop & Compact$_{\text{pre}}$
& $+0.6$ & $[-1.1,+2.3]$
& $+10.9$ & $[+7.5,+14.5]$ \\
& Traj-drop & Compact$_{\text{suf}}$
& $+2.8$ & $[+1.0,+4.6]$
& $+10.9$ & $[+7.7,+14.2]$ \\
\addlinespace[2pt]
& Compact$_{\text{pre}}$ & Raw
& $+2.5$ & $[+0.9,+4.2]$
& $+1.3$ & $[-1.7,+4.3]$ \\
& Compact$_{\text{suf}}$ & Raw
& $+0.4$ & $[-1.2,+1.9]$
& $+1.3$ & $[-1.4,+4.0]$ \\
& Compact$_{\text{pre}}$ & Compact$_{\text{suf}}$
& $+2.2$ & $[+0.7,+3.7]$
& $-0.0$ & $[-2.4,+2.4]$ \\
\midrule
\multirow{6}{*}{Downshift}
& Traj-drop & Raw
& $-4.7$ & $[-7.3,-2.1]$
& $-5.6$ & $[-8.6,-2.6]$ \\
\addlinespace[2pt]
& Traj-drop & Compact$_{\text{pre}}$
& $-5.9$ & $[-8.5,-3.4]$
& $-4.4$ & $[-7.1,-1.8]$ \\
& Traj-drop & Compact$_{\text{suf}}$
& $-2.9$ & $[-5.3,-0.4]$
& $-5.0$ & $[-7.8,-2.3]$ \\
\addlinespace[2pt]
& Compact$_{\text{pre}}$ & Raw
& $+1.2$ & $[-0.6,+3.0]$
& $-1.1$ & $[-3.2,+1.0]$ \\
& Compact$_{\text{suf}}$ & Raw
& $-1.9$ & $[-4.0,+0.3]$
& $-0.6$ & $[-2.7,+1.6]$ \\
& Compact$_{\text{pre}}$ & Compact$_{\text{suf}}$
& $+3.0$ & $[+1.1,+5.1]$
& $-0.6$ & $[-2.4,+1.3]$ \\
\bottomrule
\end{tabular}
\caption{\textbf{Pairwise interface pass-rate differences by model family.}
$\Delta$Pass is the first strategy minus the second, in percentage points.
Within each direction, rows present the central Traj-drop--Raw reversal,
broader Traj-drop--compaction comparisons, and secondary comparisons among
context-preserving interfaces, in that order. Intervals are pointwise 95\%
task-clustered bootstrap confidence intervals from $10{,}000$ resamples.}
\label{tab:sig-primary-combined}
\label{tab:sig-primary}
\label{tab:sig-gpt-primary}
\end{table*}

\section{Lost in Conversation}
\label{app:lic}
\label{app:lic_setting}
\label{app:breakdowns_lic}

\subsection{Dataset and Protocol}

\paragraph{Corpus and task families.}
We build on the sharded-instruction corpora of \emph{Lost in Conversation}
(LiC)~\citep{laban2025llmslost}, which decompose each fully specified task
into an ordered list of requirement \emph{shards}, revealed one per user turn.
We use five task families: \textbf{Code} (45 HumanEval~\citep{chen2021codex}
and 55 LiveCodeBench~\citep{jain2024livecodebench}; $n{=}100$),
\textbf{Database} (Spider~\citep{spider}; $n{=}107$), \textbf{Actions}
(the parallel category of the Berkeley Function-Calling
Leaderboard~\citep{bfcl}; $n{=}105$), \textbf{Math}
(GSM8K~\citep{gsmk8}; $n{=}103$), and \textbf{Data-to-text}
(ToTTo~\citep{totto}; $n{=}120$), for 535 tasks total. We exclude the released
summarization family because its official evaluator is an external LLM judge.
Tasks contain \(N\in[3,12]\) shards, with per-family medians of 4--7.

\paragraph{Deterministic sharded protocol.}
The original LiC protocol uses an LLM user simulator to paraphrase and schedule
shards. We instead reveal one released shard per user turn, verbatim and in
the official order, with one assistant call after every shard and no early
termination. The final user turn appends a fixed family-specific instruction
requesting the official answer format; only the final assistant response is
evaluated. Every episode therefore contains exactly \(N\) assistant turns,
and its user-turn sequence is byte-identical across models, policies, and
switch points.

\paragraph{Models and conversations.}
We evaluate the same two within-family pairs as in the coding study: Claude
Haiku~4.5/Opus~4.7 and GPT-5.6 Luna/Sol, with the same model identifiers and
inference settings as in the coding study. LiC is chat-only and uses no tools.
Assistant turns are re-fed as plain visible text without hidden reasoning
blocks, making the inherited transcript model-agnostic. Each task--policy
cell contains one episode; the schedule and evaluators are deterministic,
while model generation is not.

\paragraph{Raw handoff and switch timing.}
At switch point \(K\), the sender answers shards \(1,\ldots,K\), and the
receiver answers shards \(K+1,\ldots,N\) in the same conversation. The
receiver inherits the full visible history, including the sender's responses,
and no message announces the model change. Because episode length is fixed,
switches fire on every task. We use \textbf{Early}, \(K=1\);
\textbf{Middle}, \(K=\lfloor N/2\rfloor\); and \textbf{Late}, \(K=N-1\).

\paragraph{Evaluation and metrics.}
We use the official LiC evaluators: numeric exact match for Math, the official AST checker for Actions, execution match
against the Spider test-suite databases~\citep{zhong-etal-2020-semantic} for
Database, sandboxed reference tests for Code, and multi-reference
sacreBLEU~\citep{post-2018-call} for Data-to-text. ``Score'' is pass rate
times 100 except for Data-to-text, where it is mean per-task BLEU times 100.

QRec and CSRet use matched single-model anchors under the same sharded
schedule. Costs sum sender-priced prefix and receiver-priced suffix calls,
including cache reads and writes. In disaggregated results, QRec is omitted
whenever the absolute LC--HC anchor gap is below five points, where the
normalized denominator is unstable. This rule omits Claude Math and GPT
Database and Math. Aggregate QRec includes all five task families. 

\subsection{Results by Switch Position}

\Cref{tab:multiturn_raw_handoff_by_switch} disaggregates the Claude results
by structural switch position. Later escalation retains more savings, while
quality is mostly stable. Later downshift generally improves quality while retaining less
savings.

\Cref{tab:multiturn_raw_handoff_by_switch_gpt} reports the GPT replication.
The timing--direction interaction is similar: later escalation retains more
savings, whereas later downshift retains less. The main departure is Actions
downshift, where the small LC--HC anchor gap makes QRec volatile and the
handoff falls below LC-only at the middle and late positions.

\section{BrowseComp}
\label{app:browsecomp}
\label{app:browsecomp_setting}

\paragraph{Benchmark and task selection.}
We use BrowseComp~\citep{browsecomp}, a benchmark of 1,266 difficult
information-seeking questions with exact reference answers. To focus on tasks
that require browsing, we first sample 300 questions stratified by topic. We
remove 22 questions that a prior Claude Haiku--Opus screen answered successfully
without browsing, then select a fixed cohort of 200 questions, again stratified
by topic, from the remaining 278. Because the screening used the Claude pair,
we do not claim that the resulting cohort is specifically decontaminated for
the GPT-5.6 models evaluated here.

\paragraph{Models and browsing harness.}
As in the coding study, we use GPT-5.6 Luna and GPT-5.6 Sol as LC and HC, respectively. Each model operates in a single-context, ReAct-style loop with web-search and URL-retrieval tools, where one assistant inference counts as one step. Episodes allow up to 100 steps, followed when needed by a browsing-disabled finalization call. Research stops at approximately 80\% context utilization, preserving capacity for pending tool output and finalization.

\paragraph{Raw handoff and switch timing.}
At switch step \(K\), the receiver inherits the sender's complete
response prefix, including reasoning items, tool calls, and tool outputs.
Prefix tools are not re-executed, no message announces the change, and both
models share the original step budget. Switches fire only if the sender's
recorded trajectory continues beyond \(K\). As in the SWE setup, we calibrate sender-specific switch points using percentiles. The reported p25 and p50 points are \(K=15,31\) for escalation and
\(K=10,24\) for downshift. Each result uses the switched-only
subset.

\paragraph{Evaluation and metrics.}
Answers are graded with the official BrowseComp judge prompt and parsing rule,
using Claude Opus~4.7 as judge; grader costs are excluded. QRec and CSRet use
matched LC-only and HC-only anchors separately at each switch point and are
then averaged unweighted over p25 and p50. \emph{Full LC + full HC} pays for
both complete trajectories and uses the HC outcome. \emph{Abort + fresh HC}
pays for the LC prefix followed by a fresh complete HC trajectory.

\paragraph{Run accounting.}
Infrastructure failures are retried with backoff; exhausted episodes are
rerun from scratch and never scored partially. Each task--policy cell contains
one successfully audited generation, so we interpret these results
descriptively. Retry attempts were not retained as a separate aggregate count.





\begin{table*}
\centering
\scriptsize

\newcommand{\lictask}[2]{%
  \raisebox{-0.15em}{\includegraphics[height=1.1em]{figures/task_icons/#1.png}}\,#2}
\newcommand{\licbaseline}[5]{%
  \cellcolor{gray!12}#1 &
  \cellcolor{gray!12}#2 & \cellcolor{gray!12}#3 &
  \cellcolor{gray!12}#4 & \cellcolor{gray!12}#5 &
  \cellcolor{gray!12}#2 & \cellcolor{gray!12}#3 &
  \cellcolor{gray!12}#4 & \cellcolor{gray!12}#5 &
  \cellcolor{gray!12}#2 & \cellcolor{gray!12}#3 &
  \cellcolor{gray!12}#4 & \cellcolor{gray!12}#5}

\setlength{\tabcolsep}{1.8pt}
\renewcommand{\arraystretch}{1.03}
\resizebox{\textwidth}{!}{%
\begin{tabular}{@{}ll*{3}{rrrr}@{}}
  \toprule
  Task & Policy
    & \multicolumn{4}{c}{Early}
    & \multicolumn{4}{c}{Middle}
    & \multicolumn{4}{c}{Late} \\
  \cmidrule(lr){3-6}\cmidrule(lr){7-10}\cmidrule(l){11-14}
    & & Score & Cost & QRec & CSRet
      & Score & Cost & QRec & CSRet
      & Score & Cost & QRec & CSRet \\
    & & & (\$) & (\%) & (\%)
      & & (\$) & (\%) & (\%)
      & & (\$) & (\%) & (\%) \\
  \midrule
  \multirow{4}{*}{\lictask{python}{Code}}
    & \licbaseline{LC-only}{65.0}{0.033}{$0$}{$100$} \\
    & \licbaseline{HC-only}{92.0}{0.163}{$100$}{$0$} \\
    & Escalation
      & 88.0 & 0.145 & $85$ & $14$
      & 89.0 & 0.133 & $89$ & $23$
      & 80.0 & 0.065 & $56$ & $75$ \\
    & Downshift
      & 75.0 & 0.047 & $37$ & $89$
      & 85.0 & 0.082 & $74$ & $62$
      & 85.0 & 0.156 & $74$ & $5$ \\
  \midrule
  \multirow{4}{*}{\lictask{database}{Database}}
    & \licbaseline{LC-only}{51.4}{0.011}{$0$}{$100$} \\
    & \licbaseline{HC-only}{72.0}{0.061}{$100$}{$0$} \\
    & Escalation
      & 67.3 & 0.068 & $77$ & $-12$
      & 65.4 & 0.049 & $68$ & $25$
      & 67.3 & 0.024 & $77$ & $73$ \\
    & Downshift
      & 50.5 & 0.019 & $-5$ & $85$
      & 50.5 & 0.032 & $-5$ & $58$
      & 55.1 & 0.050 & $18$ & $22$ \\
  \midrule
  \multirow{4}{*}{\lictask{apis}{Actions}}
    & \licbaseline{LC-only}{57.1}{0.006}{$0$}{$100$} \\
    & \licbaseline{HC-only}{70.5}{0.035}{$100$}{$0$} \\
    & Escalation
      & 73.3 & 0.032 & $121$ & $10$
      & 69.5 & 0.024 & $93$ & $38$
      & 70.5 & 0.013 & $100$ & $77$ \\
    & Downshift
      & 59.0 & 0.010 & $14$ & $85$
      & 61.0 & 0.018 & $29$ & $58$
      & 61.0 & 0.029 & $29$ & $22$ \\
  \midrule
  \multirow{4}{*}{\lictask{math}{Math}}
    & \licbaseline{LC-only}{90.3}{0.008}{--}{$100$} \\
    & \licbaseline{HC-only}{94.2}{0.052}{--}{$0$} \\
    & Escalation
      & 94.2 & 0.049 & -- & $6$
      & 94.2 & 0.040 & -- & $27$
      & 94.2 & 0.018 & -- & $77$ \\
    & Downshift
      & 89.3 & 0.013 & -- & $90$
      & 91.3 & 0.023 & -- & $66$
      & 94.2 & 0.043 & -- & $20$ \\
  \midrule
  \multirow{4}{*}{\lictask{data2text}{Data-to-text}}
    & \licbaseline{LC-only}{46.5}{0.015}{$0$}{$100$} \\
    & \licbaseline{HC-only}{55.6}{0.089}{$100$}{$0$} \\
    & Escalation
      & 55.4 & 0.081 & $97$ & $10$
      & 52.9 & 0.061 & $70$ & $37$
      & 51.9 & 0.041 & $59$ & $65$ \\
    & Downshift
      & 47.4 & 0.028 & $9$ & $83$
      & 49.9 & 0.061 & $37$ & $38$
      & 51.6 & 0.076 & $56$ & $16$ \\
  \bottomrule
\end{tabular}%
}
\caption{\textbf{Raw handoff by switch position in sharded, underspecified
conversations, Claude pair.}
Early, middle, and late are structural switch positions. Later escalation
retains more savings, while quality is mostly stable unless prior turns
accumulate deliverable state. Later downshift generally improves quality while
retaining less savings. Score is pass rate except for Data-to-text, which
reports BLEU$\times 100$.}
\label{tab:multiturn_raw_handoff_by_switch}
\end{table*}

  \begin{table*}
  \centering
  \scriptsize

  \providecommand{\lictask}[2]{%
    \raisebox{-0.15em}{\includegraphics[height=1.1em]{figures/task_icons/#1.png}}\,#2}
  \providecommand{\licbaseline}[5]{%
    \cellcolor{gray!12}#1 &
    \cellcolor{gray!12}#2 & \cellcolor{gray!12}#3 &
    \cellcolor{gray!12}#4 & \cellcolor{gray!12}#5 &
    \cellcolor{gray!12}#2 & \cellcolor{gray!12}#3 &
    \cellcolor{gray!12}#4 & \cellcolor{gray!12}#5 &
    \cellcolor{gray!12}#2 & \cellcolor{gray!12}#3 &
    \cellcolor{gray!12}#4 & \cellcolor{gray!12}#5}

  \setlength{\tabcolsep}{1.8pt}
  \renewcommand{\arraystretch}{1.03}
  \resizebox{\textwidth}{!}{%
  \begin{tabular}{@{}ll*{3}{rrrr}@{}}
    \toprule
    Task & Policy
      & \multicolumn{4}{c}{Early}
      & \multicolumn{4}{c}{Middle}
      & \multicolumn{4}{c}{Late} \\
    \cmidrule(lr){3-6}\cmidrule(lr){7-10}\cmidrule(l){11-14}
      & & Score & Cost & QRec & CSRet
        & Score & Cost & QRec & CSRet
        & Score & Cost & QRec & CSRet \\
      & & & (\$) & (\%) & (\%)
        & & (\$) & (\%) & (\%)
        & & (\$) & (\%) & (\%) \\
    \midrule
    \multirow{4}{*}{\lictask{python}{Code}}
      & \licbaseline{LC-only}{85.0}{0.017}{$0$}{$100$} \\
      & \licbaseline{HC-only}{95.0}{0.057}{$100$}{$0$} \\
      & Escalation
        & 96.0 & 0.052 & $110$ & $13$
        & 94.0 & 0.037 & $90$ & $49$
        & 94.0 & 0.024 & $90$ & $82$ \\
      & Downshift
        & 90.0 & 0.024 & $50$ & $82$
        & 94.0 & 0.037 & $90$ & $50$
        & 96.0 & 0.052 & $110$ & $13$ \\
    \midrule
    \multirow{4}{*}{\lictask{database}{Database}}
      & \licbaseline{LC-only}{65.4}{0.008}{--}{$100$} \\
      & \licbaseline{HC-only}{61.7}{0.027}{--}{$0$} \\
      & Escalation
        & 64.5 & 0.025 & -- & $14$
        & 65.4 & 0.019 & -- & $44$
        & 66.4 & 0.014 & -- & $70$ \\
      & Downshift
        & 60.7 & 0.012 & -- & $80$
        & 59.8 & 0.017 & -- & $52$
        & 64.5 & 0.024 & -- & $18$ \\
    \midrule
    \multirow{4}{*}{\lictask{apis}{Actions}}
      & \licbaseline{LC-only}{73.3}{0.005}{$0$}{$100$} \\
      & \licbaseline{HC-only}{79.0}{0.028}{$100$}{$0$} \\
      & Escalation
        & 82.9 & 0.023 & $167$ & $23$
        & 79.0 & 0.017 & $100$ & $50$
        & 82.9 & 0.009 & $167$ & $83$ \\
      & Downshift
        & 73.3 & 0.011 & $0$ & $76$
        & 68.6 & 0.017 & $-83$ & $49$
        & 70.5 & 0.023 & $-50$ & $23$ \\
    \midrule
    \multirow{4}{*}{\lictask{math}{Math}}
      & \licbaseline{LC-only}{92.2}{0.005}{--}{$100$} \\
      & \licbaseline{HC-only}{94.2}{0.024}{--}{$0$} \\
      & Escalation
        & 94.2 & 0.022 & -- & $10$
        & 92.2 & 0.018 & -- & $32$
        & 96.1 & 0.010 & -- & $76$ \\
      & Downshift
        & 89.3 & 0.007 & -- & $91$
        & 91.3 & 0.011 & -- & $68$
        & 91.3 & 0.019 & -- & $24$ \\
    \midrule
    \multirow{4}{*}{\lictask{data2text}{Data-to-text}}
      & \licbaseline{LC-only}{33.4}{0.008}{$0$}{$100$} \\
      & \licbaseline{HC-only}{49.9}{0.036}{$100$}{$0$} \\
      & Escalation
        & 49.3 & 0.029 & $96$ & $24$
        & 49.5 & 0.024 & $98$ & $40$
        & 45.6 & 0.018 & $74$ & $63$ \\
      & Downshift
        & 32.9 & 0.009 & $-3$ & $95$
        & 32.7 & 0.023 & $-4$ & $45$
        & 37.0 & 0.030 & $22$ & $22$ \\
    \bottomrule
  \end{tabular}%
  }
  \caption{\textbf{Raw handoff by switch position in sharded, underspecified
  conversations, GPT pair.}
  Early, middle, and late are structural switch positions. Later escalation
  retains more savings, while quality is mostly stable unless prior turns
  accumulate deliverable state. Later downshift generally improves quality while
  retaining less savings. Score is pass rate except for Data-to-text, which
  reports BLEU$\times 100$.}
  \label{tab:multiturn_raw_handoff_by_switch_gpt}
  \end{table*}


\begin{table*}
  \centering
  \small
  \renewcommand{\arraystretch}{1.12}
  \begin{tabular}{@{}llrrrrrrrr@{}}
    \toprule
    \multirow{2}{*}{Prefix model}
      & \multirow{2}{*}{Difficulty}
      & \multirow{2}{*}{$N$}
      & \multicolumn{7}{c}{Switch step by prefix-run percentile} \\
    \cmidrule(lr){4-10}
      & & & p5 & p10 & p15 & p25 & p35 & p45 & p50 \\
    \midrule

    \multirow{3}{*}{Haiku~4.5}
      & Easy   & 194 & 25 & 30 & 34 & 37 & 42 & 46 & 50 \\
      & Medium & 261 & 35 & 41 & 45 & 51 & 58 & 63 & 66 \\
      & Hard   & 45  & 48 & 54 & 58 & 63 & 77 & 86 & 94 \\
    \midrule

    \multirow{3}{*}{Opus~4.7}
      & Easy   & 194 & 5  & 7  & 8  & 10 & 12 & 14 & 14 \\
      & Medium & 261 & 8  & 12 & 13 & 16 & 18 & 21 & 23 \\
      & Hard   & 45  & 18 & 19 & 20 & 24 & 30 & 33 & 37 \\
    \midrule

    \multirow{3}{*}{GPT-5.6 Luna}
      & Easy   & 194 & 7 & 8  & 8  & 9  & 9  & 10 & 10 \\
      & Medium & 261 & 8 & 8  & 9  & 10 & 10 & 11 & 11 \\
      & Hard   & 45  & 9 & 10 & 11 & 11 & 11 & 12 & 12 \\
    \midrule

    \multirow{3}{*}{GPT-5.6 Sol}
      & Easy   & 194 & 9  & 10 & 10 & 11 & 12 & 13 & 13 \\
      & Medium & 261 & 10 & 11 & 11 & 13 & 13 & 15 & 15 \\
      & Hard   & 45  & 12 & 13 & 14 & 15 & 16 & 18 & 19 \\
    \bottomrule
  \end{tabular}

  \caption{\textbf{Difficulty-calibrated switch steps by prefix model.}
  Steps are agent turns, corresponding to model API calls. Each threshold is
  the specified percentile of the prefix model's single-model termination-step
  distribution within the given difficulty bucket.}
  \label{tab:step-distribution}
\end{table*}

\end{document}